\RequirePackage{fix-cm}

\documentclass[twocolumn]{svjour3} 

\usepackage{graphicx}
\usepackage{verbatim}
\usepackage[english]{babel}
\usepackage[utf8]{inputenc}
\usepackage[T1]{fontenc}
\usepackage{pdfpages}
\usepackage{time}
\usepackage{appendix}
\usepackage{sidecap}
\usepackage{subfigure}

\usepackage{chngcntr}

\usepackage{url}
\usepackage[]{algorithm2e}
\usepackage{rotating}
\usepackage{multirow}
\usepackage{listings}
\usepackage{amsmath}
\usepackage{amsfonts}
\usepackage{amssymb}
\usepackage{hyphenat}
\usepackage[bottom]{footmisc}
\usepackage{array}
\usepackage[misc]{ifsym}
\usepackage{pgfplots}
\usepackage{graphicx}

\usepackage{hyperref}
\usepackage{textcomp}
\usepackage{tabularx}
\usepackage{makecell}

\usepgfplotslibrary{statistics}
\usetikzlibrary{patterns}

\usepackage{comment}
\usepackage{pgfplotstable}
\usepackage{float}
\usepackage[acronym]{glossaries}
\usepackage{enumitem}

\begin{document}

\title{Monocular Trail Detection and Tracking Aided by Visual SLAM for Small Unmanned Aerial Vehicles}
\author{Andr\'e Silva \and Ricardo Mendon\c ca \and Pedro Santana}
\authorrunning{A. Silva, R. Mendon\c ca, P. Santana} 
\institute{A. Silva \at Instituto de Telecomunica\c c\~oes, Lisboa, Portugal 
\\ ISCTE-Instituto Universit\'ario de Lisboa, Lisboa, Portugal 
\\ \email{afgsa1@iscte-iul.pt} \and R. Mendonça \at CTS-UNINOVA, Monte de Caparica, Portugal 
\\  \email{rmm@uninova.pt} \and \Letter~P. Santana \at Instituto de Telecomunica\c c\~oes, Lisboa, Portugal 
\\ ISCTE-Instituto Universit\'ario de Lisboa, Lisboa, Portugal 
\\ \email{pedro.santana@iscte-iul.pt} 
}

\date{Received: date / Accepted: date}

\maketitle

\begin{abstract}
This paper presents a monocular vision system susceptible of being installed in unmanned small and medium-sized aerial vehicles built to perform missions in forest environments (e.g., search and rescue). The proposed system extends a previous monocular-based technique for trail detection and tracking so as to take into account volumetric data acquired from a Visual SLAM algorithm and, as a result, to increase its sturdiness upon challenging trails. The experimental results, obtained via a set of 12 videos recorded with a camera installed in a tele-operated, unmanned small-sized aerial vehicle, show the ability of the proposed system to overcome some of the difficulties of the original detector, attaining a success rate of $97.8\,\%$.
\end{abstract}


\section{Introduction}

\begin{figure}[t]
  \centering
  \includegraphics[height=2.5cm]{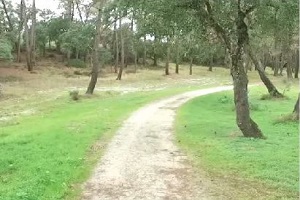}\label{fig:f1_1}
\includegraphics[height=2.5cm]{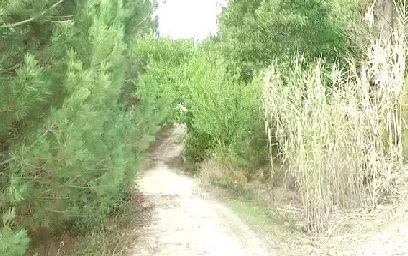}\label{fig:f1_2}
  \caption{Typical trails.}\label{fig:fig1}
\end{figure}

Aerial platforms become particularly interesting for use in environments where land-based platforms have more difficulties to access, such as forests. Forests are often populated with a dense and extensive network of trails. These trails are, by definition, considered to be ground communication routes that are usually free of obstacles and, thus, allow people and vehicles to traverse complex environments more efficiently. In this sense, an Unmanned Aerial Vehicle (UAV) able to follow these trails should be able to reduce the computational cost associated with trajectory (re)planning, as there are fewer obstacles (e.g., trees) in the path of the vehicle to avoid. The ability of following trails also allow UAV to fly low so as to obtain ground-level detailed imagery, which would be otherwise concealed to satellite and high-altitude aerial vehicles.

Trails do not exhibit the typical markings of roads nor are paved structures (see Fig.~\ref{fig:fig1}) and, thus, the known techniques for road detection (e.g., \cite{sadano2002lane,hillel2014recent,kenue1988detection}) cannot be applied to trail detection. These challenges led to the development of a set of specific techniques for path and trail navigation based on computer vision \cite{bartel2007real,rasmussen2008shape,rasmussen2004grouping,rasmussen2014trail,santana2010saliency,santana2011swarms,santana2013neural}, which respond to the needs of safe navigation in natural environments where GPS navigation is insufficient \cite{zheng2005quantitative}, as it is often the case when flying below canopy level. When compared to alternative sensory modalities (e.g., laser scanners), vision systems have fewer requirements in terms of payload and energetic capacity, which is key for small sized UAV with small energetic autonomy.

This paper contributes to push forward vision-based trail detection in forest environments in the context of small UAV. To this end, this work extends a prior saliency-based monocular trail detection and tracking technique \cite{santana2013neural} in order to increase its robustness in the presence of more challenging trails. Although this previous technique showed to be robust in a diverse and challenging dataset, it failed to deliver accurate results in the presence of strong distractors in the environment. These distractors have a significant impact in this previous technique as it relies on tracking by learning, meaning that once it fails to track the trail (due to the presence of a distractor) only hardly it will be able to recover from the failure. 

To reduce the chances of being mistakenly attracted by distractors, this paper extends the previous technique \cite{santana2013neural} so as to take into account the volumetric structure of the environment, using visual SLAM for the purpose. The use of visual SLAM over the obvious alternative stereo-vision aims at fostering future migration of the system to commercial platforms, which are most equipped with monocular vision systems. Alongside the extended detector and tracker, this paper proposes a mechanism to determine in runtime whether the detector is performing correctly or not. This is crucial to help the navigation system to determine if the detector is appropriate for the current context.
 
To allow an easy integration on existing platforms, the proposed system was implemented on top of the Robot Operating System (ROS) \cite{288}. The system hereby proposed was tested using a 12 videos dataset, composed of 128870 frames, acquired on field runs using a small commercial UAV. The obtained results show an increase in robustness when compared to the original system, obtaining an overall $97.8\,\%$ success rate on the tested dataset.

This article is organized as follows. Section~\ref{cap:stateofart} briefly introduces the State of Art of current technologies regarding path following. Section~\ref{cap:implementation} details the work developed during the implementation of the purposed system in this article. Section~\ref{cap:results} presents the results obtained with the proposed system in comparison with the work previously done by Santana et. al. \cite{santana2013neural}. Section~\ref{cap:cfw} resumes the results achieved throughout this work and presents some future improvements to yield better results.

\section{Related Work}\label{cap:stateofart}

Current path detection methods rely considerably on work developed for ill-structured unpaved rural roads. The typical solution in this case is to segment the road region from its surroundings by considering a set of pixels whose the probability of belonging to the road surface is above a given threshold calculated through models learned off-line \cite{giusti2016machine,goodrich2008supporting} or on-line in a self-supervised way \cite{pp1960automatic,itti2000saliency}. Frequently, a simplified known model of the road (e.g., trapezoidal) is fit to the segmented images. The Region Growing technique is an alternative to the model fitting process for less structured roads \cite{kenue1988detection,holz2011real,giusti2016machine}. By enforcing a global shape constraint, the model-based approach enables the substitution of the road/non-road pixel classification process by an unsupervised clustering mechanism \cite{klasing2009comparison}.

The models referred in the previous paragraph are the basis of most work on path detection. An example is the use of \textit{a priori} knowledge about the color distributions of both the path and their surroundings for segmentation \cite{langelaan2005towards}. The main direction of the path can also be learned off-line in a supervised way \cite{giusti2016machine}. The use of off-line learned models has the limitation that the robot is only able to perceive environments that have been covered by the training set. Robustness can be increased if these \textit{a priori} models are substituted by models learned on-line \cite{luccheseyz2001color,malik2001contour}. In contrast to the road domain, the definition of the reference regions from which it is possible to supervise the learning process is challenging. With varying width and orientation, it is difficult to assure that the robot is on the path, and from that, which regions of the input image can be used as reference. Moreover, paths and its surroundings often exhibit the same height, which hampers a straightforward use of depth information to determine a path reference patch. 

The use of a global shape constraint (e.g., triangular) to avoid the learning process has also been tested in the path detection domain \cite{mccall2006video}. This is done by over-segmenting the image, creating sets of segments, and then scoring them against the global shape constraint. Accurate image over-segmentation is a computationally demanding task and usually requires clear edges segmenting the object from the background. Moreover, global shape constraints limit the type of paths that can be detected.

To overcome the previous work's limitations, Rasmussen et al. \cite{rasmussen2004grouping} proposed the use of local appearance contrast for path detection. Later, Santana et al. \cite{santana2013neural} exploited the related concept of visual saliency to detect paths. Visual saliency was exploited with a swarm-based solution, inspired by the social insects metaphor \cite{bonabeau1999swarm}, capable of interpreting the input images without the cost of explicitly over-segmenting the input image and the brittleness of depending on accurate appearance and shape models. As mentioned before, this method has shown to not exhibit sufficient robustness in the presence of strong distractors in the environment. This paper proposes an extension to the this previous method, based on 3-D data acquired from a visual SLAM technique, to reduce the chances of the method to be mistakenly attracted by distractors in the environment. 

The use of three-dimensional information to improve the targeting methods for detection of trails was initially explored by Rasmussen et al. \cite{rasmussen2011integrating,rasmussen2014trail}. More specifically, Rasmussen et al. \cite{rasmussen2014trail} used a binocular vision system composed of two omni-directional cameras and a laser scanner to map the vehicle's surrounding area. Thus, this previous method also presents itself as a solid solution for trail segmentation but requires the use of a laser scanner, which should be avoided if size, weight, and energy constraints found in small UAV are to be met. Fleischmann et al. \cite{fleischmann2016adaptive} also proposed a system based on binocular vision for detection and tracking of paths in a forest environment. In this paper we offer a solution that is based on monocular vision and, thus, is more easily integrated in current commercial UAV solutions. Moreover, the ability to solve the trail detection problem with a single camera is key for robots equipped with binocular systems when one of the camera fails or when stereo calibration data becomes deprecated.

\section{Proposed System}\label{cap:implementation}

\subsection{System overview}\label{sec:overview}

The proposed system, presented briefly in this section and detailed in the following, is based on the path detector proposed by Santana et al. \cite{santana2013neural}, hereinafter referred to as \textit{original detector}. The original detector is able to exploit the fact that trails are  salient in the images acquired by the robot's on-board camera to learn on-line the appearance of the trail. This learned appearance is used to adjust the detection process and a temporal filter is employed to improve the detector's robustness. With these characteristics, the original detector is robust in the presence of a large set of poorly structured trails. However, the presence of distracting elements in their visual field, i.e., elements that exhibit an higher visual saliency than the trail itself, such as tree trunks and certain types of dense vegetation, may elude the original detector, leading it to misidentify these regions as belonging to the trail. For a robust operation, the robot must be able to avoid these situations and, when this is not possible, it must be able to identify the fault occurrence so that its control system reacts appropriately. 

In order to reduce the original detector's failure rate, this work proposes the use of 3-D information to inform the detector about the presence of objects that can not be considered as belonging to the trail, like trees and bushes. In this way, the original detector should be able to avoid considering the regions of the image where these objects lie, focusing its operation on the remaining regions, thus reducing the probability of producing erroneous results due to the presence of distractors in the robot's visual field. To keep the system compact and flexible, i.e., to maintain the system's ability to be mounted on robots with weight constraints, such as small unmanned aerial vehicles, a monocular vision system is used for the acquisition of 3-D information. 

The original detector depends on a set of parameters that need empirical tuning. The system herein proposed  seeks to show that the effort required for this parameterization can be reduced by making one of these parameters dependent on the camera's orientation, as estimated based on the environment's 3-D reconstruction.

\begin{figure}
  \centering
  \includegraphics[width=7cm]{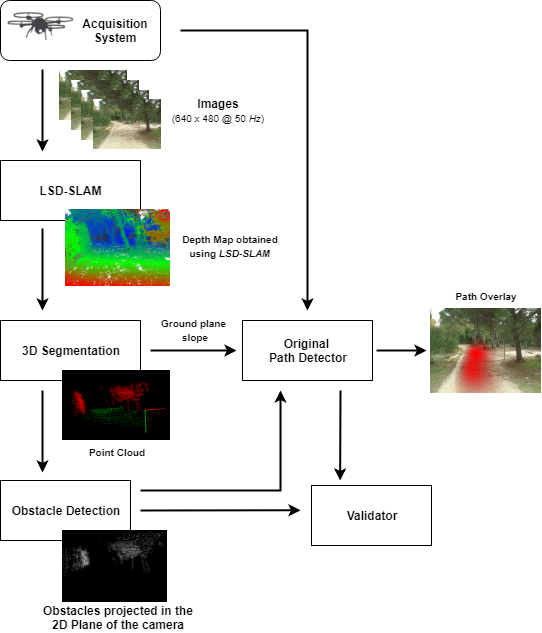}
  \caption{System overview.}\label{fig:system_overview}
\end{figure}

Fig.~\ref{fig:system_overview} presents a block diagram of the proposed system, organized in four main modules. The first module (detailed in Section~\ref{sec:acquisition}) is responsible for the environment's 3-D reconstruction, using a vision-based Simultaneous Localization and Mapping (SLAM) technique, namely LSD-SLAM~\footnote{Large-Scale Direct Monocular SLAM} \cite{engel2014lsd}. With LSD-SLAM, the robot is able to obtain a set of depth maps from the video stream captured with its on-board monocular camera. These depth maps are then used to obtain 3-D point clouds whose coordinates are described in the camera's frame of reference.

The second module (detailed in Section~\ref{sec:segmentation}) comprises a set of processes used to segment and analyze the information contained in the obtained point clouds: first, the ground surface is estimated assuming it can be locally modeled as a plane in the frontal region of the robot; then, all points above the ground plane are classified as possible obstacles, i.e., distracting elements to the original trail detector (e.g., trees and high vegetation). The 3-D points of each obstacle are then projected into the camera plane to produce a 2-D mask that signals the regions of the image that contain potential distracting elements. As previously stated, the detector also receives information concerning the camera inclination, estimated from the relative inclination of the ground plane.

The third module (detailed in Section~\ref{sec:integration}) uses the data calculated in the previous section, i.e., the 2-D mask signaling the obstacle regions and the inclination value of the plane in relation to the ground plane, to influence the behavior of the original detector. Finally, the fourth module (detailed in Section~\ref{sec:validationsystem}) describes the validation process of the proposed system, where the results obtained by the trail detector are analyzed taking into account the obstacle regions previously identified, that is, the cases where there is an interception between the region hypothesis of the trail and the regions characterized as obstacle are verified. This information can then be made available to the robot's high-level control system as a degree of certainty that the trail location has been correctly identified.

\subsubsection{Original Path Detector}\label{sec:pathdetector}

\begin{figure}
  \centering
  {\includegraphics[width=8cm]{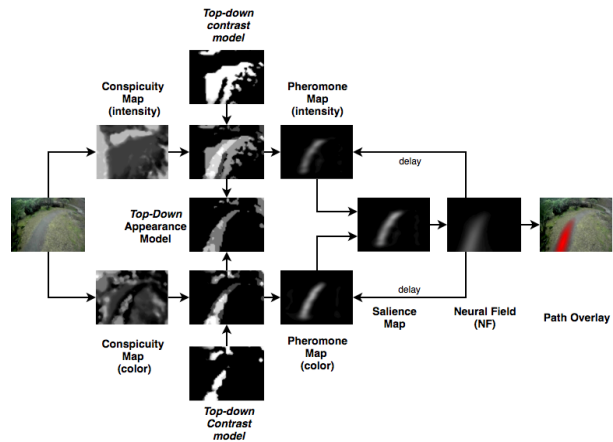}
  \hfill
  \caption{Overview of the \textit{original detector} \cite{santana2013neural}. Image adapted from \cite{santana2013neural}.}\label{fig:pathd}}
\end{figure}

As mentioned, paths are expected to appear in the robot's visual field as salient structures in the environment. Using this feature, Santana et al. \cite{santana2013tracking,santana2013neural} proposed a method that exploits visual saliency models for the purpose of path detection (see Fig.~\ref{fig:pathd}). The following describes the method in a nutshell, focusing on the relevant aspects influencing the extensions herein proposed. A detailed description can be found elsewhere \cite{santana2013tracking,santana2013neural}. 

The model operates on video streams of color images, producing an estimate of path location at each frame. The model's first step, in each frame, is to compute feature-dependent conspicuity maps, based on the well-known bio-inspired model of Itti et al. \cite{itti2000saliency}. As a result, a color  conspicuity map and an intensity conspicuity map are created based on a set of center-surround feature maps at various spatial scales. These maps identify the regions of the robot's visual field that detach more from the background on the visual feature in question and, hopefully, most likely to belong to the path.

\begin{figure}
  \centering
  \subfigure[]{\includegraphics[height=3.2cm]{figures/ant.png}
  \label{fig:f10_0}}
  \subfigure[]{\includegraphics[height=3.2cm]{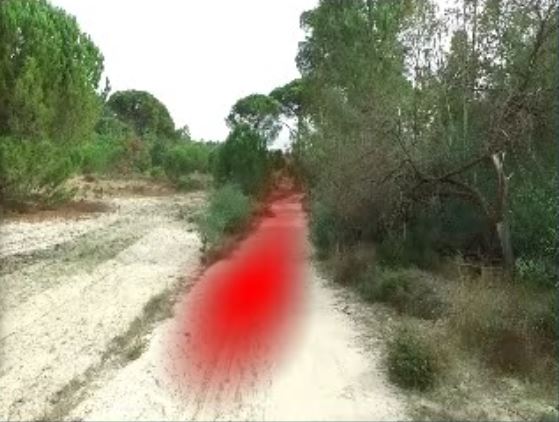}
  \label{fig:f10_1}}
  \hfill
  \caption{Original path detector in operation. Left: possible trajectory produced by a virtual agent (black dashed line) up to an empirically defined threshold $h_{max}$ (red dashed line). Right: a typical result obtained with the original detector when operating correctly.}\label{fig:ant}
\end{figure}

Both conspicuity maps are used as sensory input for two sets of virtual agents (see Fig.~\ref{fig:ant}) that collectively create a color pheromone map and an intensity pheromone map, based on the ant foraging metaphor \cite{dorigo2011ant}. These pheromone maps represent the self-organized consensus reached by the agents regarding the path's skeleton. Iterated several times, each agent moves on the conspicuity map on which it was deployed and deposits a given amount of virtual pheromone along is trajectory in the corresponding pheromone map. This pheromone is intended to indirectly influence the behavior of the other agents towards a consensus. 

Virtual agents are controlled by a set of behaviors that vote for the next agents move (e.g., move one pixel up, move one pixel to the right) according to local distribution of pheromone, on-line self-supervised learned path appearance/saliency, and \textit{a priori} knowledge regarding paths typical overall structure. The local distribution of pheromone aims at allowing agents to indirectly interact and, as a result, self-organize. The on-line learned appearance/saliency helps agents to focus their activity in regions that are salient in the visual field and best resemble the appearance of the path in the previous frames. The \textit{a priori} knowledge of path's structure aims at biasing agents to move along a trajectory that matches some expectations regarding the path's global shape without hampering them from exploiting the sensory/pheromone clues. Examples of these biases include inertia under the assumption that paths' orientation tend to be monotonous and tendency to move towards zones equidistant to the edges in the conspicuity map closest to the agent. With these behaviors, agents move from the bottom region of the input image up to a maximum number of steps or a given maximum height (an image row) (see Fig.~\ref{fig:f10_0}). The maximum height $h_{max}$ is empirically defined in order to avoid agents to analyze regions above the horizon line.

Then, both pheromone maps are combined to produce a single saliency map. In order to improve robustness, pheromone is accumulated across frames by feeding a dynamical neural field with the frame-wise saliency maps, where $F_i(x,y)\in [0,1]$ provides the activity level of the neural field at pixel $(x,y)$ at frame $i$. The dynamic of the neural field implements pheromone evaporation and spatial competition, helping the system to converge to a solution in a robust way. To reduce the coupling between the dynamics of the neural field and the dynamics of the robot, the homography matrix that describes the projective transformation between the current and the previous frames is estimated using optical flow cues and used to motion compensate the neural field. At each frame, the path location is given by the biggest blob in the neural field that exhibits and activity level above a given threshold (see Fig.~\ref{fig:f10_1}). 

The original path detector showed to be robust and able to identify a wide variety of paths under different conditions (e.g., see Fig.~\ref{fig:ant}). However, being primarily based on visual saliency as a clue for path presence, the presence of distractors, that is, elements that are themselves protruding in the environment, like shadows or trees, may induce the system to fail in challenging conditions. The following sections describe the extensions herein proposed to the original path detector so as to circumvent these limitations.

\subsection{Point Cloud Acquisition}\label{sec:acquisition}

The 3D reconstruction of the environment is obtained with a software package \cite{engel2014lsd} that implements the LSD-SLAM method \cite {engel2013semi}, which is able to operate with monocular cameras. Other similar solutions are available for the acquisition of 3D information based on SLAM techniques, such as ORB-SLAM \cite{mur2015orb} and REMODE \cite{pizzoli2014remode}. However, REMODE depends on the use of an Graphics Processing Unit (GPU), which is not applicable to most of commercial UAVs available and ORB-SLAM provides only a sparse representation of the scene. 

Handling every frame of a video feed at 50 frames per second can be computational intensive. To circumvent this problem, LSD-SLAM uses frames spaced in time, known as key-frames, containing the necessary information needed to infer the camera movement and the 3D reconstruction of the environment using a SLAM technique. For each key-frame $i$, LSD-SLAM provides a corresponding $w \times h$ RGB-D image, $I_i$, whose pixels include color and depth information, alongside the camera's pose in world coordinates, $S_i$, represented as an homogeneous rigid transformation matrix, where $w$ and $h$ are the number of columns and rows, respectively. The depth information and camera's intrinsic parameters, estimated during an off-line calibration process under the pinhole model assumption, are used to create a colored 3D point cloud $P_i^L$, described in camera's local frame coordinates. A colored 3D point cloud is a set of 3D points associated with color information in the RGB color space.

\begin{figure}
  \centering
  \subfigure[]{\includegraphics[width=4cm, height=2.6cm]{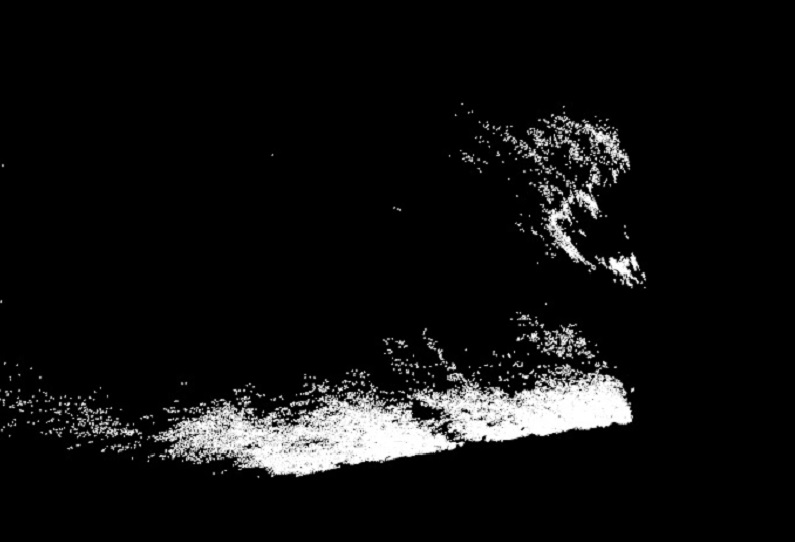}\label{fig:pointcloudresult_1}}\hfill
  \subfigure[]{\includegraphics[width=4cm, height=2.6cm]{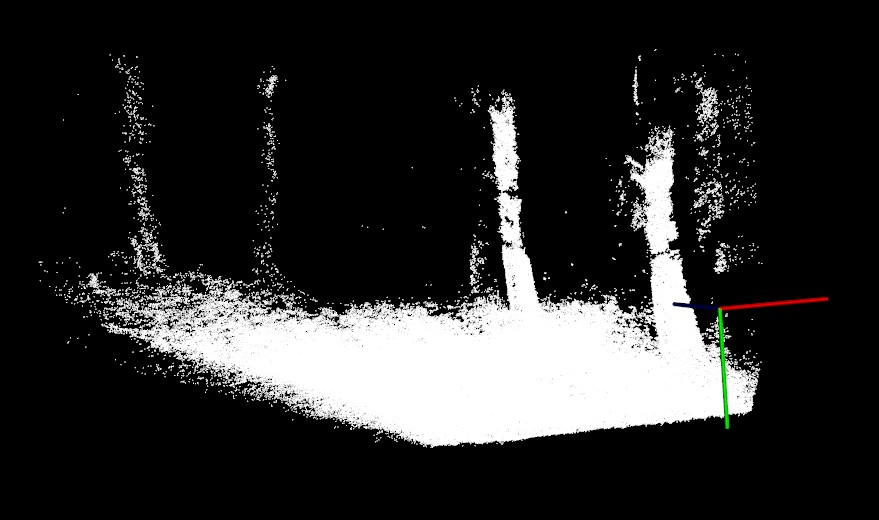}\label{fig:pointcloudresult_3}}\hfill
  \caption{Typical point clouds acquired with the proposed system. (a) A point cloud in the camera's local frame. (b) A set of ten point clouds accumulated in the world frame.}\label{fig:pointcloudresult}
\end{figure}

The point cloud produced at each key-frame provides a representation of the environment that is sparse and incomplete (see Fig.~\ref{fig:pointcloudresult}). To obtained a denser representation of the environment, point clouds acquired in the set of $n$ previous key-frames are accumulated into a single point cloud $A_i$ described in world coordinates:

\begin{equation}
A^W_i = P_{i-n} \cup \dots\cup P_{i}
\end{equation}

\noindent where $P_k$ corresponds to point cloud $P_k^L$ transformed to world coordinates with camera's pose $\mbox{S}_k$, 

\begin{equation}
P_k=\{\mbox{S}_k \mathbf{p} : \forall \mathbf{p} \in  P_k^L\}.
\end{equation}

\subsection{Point Cloud Processing}\label{sec:segmentation}

\begin{figure}
  \centering
  {\includegraphics[width=8cm]{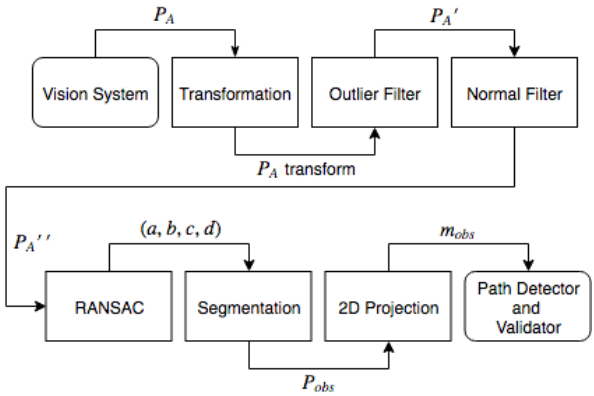}}\hfill
  \caption{Overview of the processes used to process the acquired accumulated point clouds.}\label{fig:segprocess}
\end{figure}

This section describes the processes applied to the accumulated point cloud at a given key-frame $i$, $A_i^W$ (see Fig.~\ref{fig:segprocess}). The first step consists of transforming point cloud from world coordinates to local camera coordinates using the inverse of the camera's estimated pose, $\mbox{S}_i^{-1}$:

\begin{equation}
A_i=\{\mbox{S}^{-1}_i \mathbf{p} : \forall \mathbf{p} \in  A_i^W\}.
\end{equation}

The accumulated point cloud $A_i$ can be composed of thousands of points, which makes its analysis an onerous task. As a way of circumventing this problem, a set of filters are applied to decrease its density without disrupting its structure and also to remove outliers resulting from noisy processes. Outliers are removed with a statistical method \cite{Rusu_ICRA2011_PCL} over a kd-tree computed from the point cloud \cite{silpa2008optimised} for performance purposes. The set of points in $A_i$ not considered to be outliers is represented as $A'_i, A'_i \subseteq A_i$. 

The following step concerns the detection of a dominant ground plane, which is assumed to exist in the frontal region of the robot. To reduce the chances of failing in the detection of the ground plane, points laying on surfaces unlikely to be horizontal planes are discarded (e.g., like the surface of a tree trunk). This is done by using only the points from $A_i'$ that have normal orientations pointing up by a given minimum angle, i.e., the normals perpendicular to the Y axis assumed to contain the ground plane in front of the robot. To achieve this, a technique know as Difference of Normals (DoN) \cite{ioannou2012difference} is used, where the set of points in $A'_i$ exhibiting unusual normal vectors are not taken into account in the calculation of the ground plane, thus obtaining a new set of point represented as $A''_i, A''_i \subseteq A'_i$ where the points exhibiting unusual normal are excluded. 

\subsubsection{Ground Plane and Obstacles Segmentation} \label{sec:planeestimation}

\begin{figure}
  \centering
  \subfigure{\includegraphics[width=8cm]{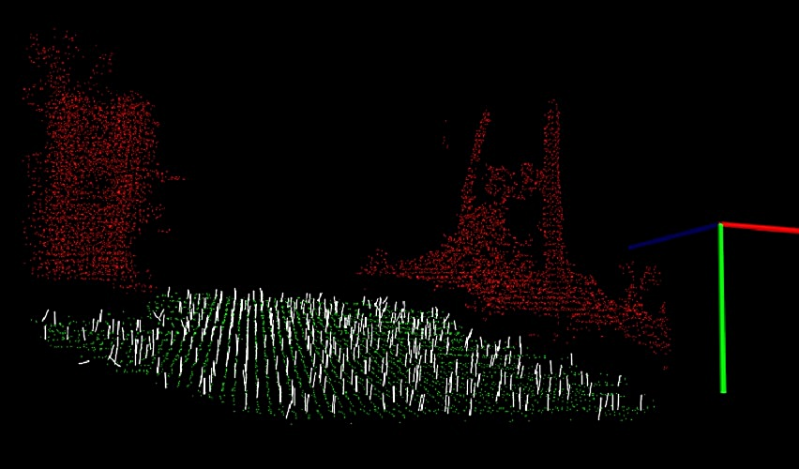}\label{fig:normal_1}}\hfill
  \caption{Ground segmentation process  based on normal vectors, green and red represent on-plane points and (plane inliers) off-plane points (plane outliers), respectively. The white vectors represent normal vectors computed for on-plane points.}\label{fig:normal}
\end{figure}

After applying the aforementioned filters, it is necessary to detect the obstacles present in the local environment, assumed to be identifiable as disturbances in the environment at the ground level. A RANSAC-based method \cite{fischler1987random} is used to find the plane $\mathbf{\pi}_i=(a_ix+b_iy+c_iz+d_i)$ with highest number of inliers associated to normal vectors (computed in the point's neighborhood) that are close to perpendicular with respect to $\mathbf{\pi}_i$ \cite{holz2011real}. The set of inliers of $\mathbf{\pi}_i$, $I_i$, is the sub-set of $A''_i, I_i \subseteq A''_i$ whose points have a perpendicular distance to plane $\mathbf{\pi}_i$  below the average of the distances between the ground plane and all the points belonging to $I_i$, $\gamma$ (see Fig.\ref{fig:normal}): 

\begin{equation}
I_i = \{\mathbf{p}\in A''_i, d(\mathbf{p},\mathbf{\pi}_i) < \gamma\}, 
\end{equation}

\begin{equation}
d(\mathbf{p},\mathbf{\pi}_i) = \frac{a_ix_p+b_iy_p+c_iz_p+d_i}{\sqrt{a_i^2+b_i^2+c_i^2}}, \mathbf{p}=(x_p,y_p,z_p).
\end{equation}

Once the ground plane $\mathbf{\pi}_i$ is found, the next step is to determine which points of $A''_i$ are sufficiently above the plane so that can be labeled as obstacles with confidence. To the set of obstacle points we call $O_i, O_i\subseteq A''_i$ (see Fig.~\ref{fig:segprocess}). To accommodate the point cloud's unknown noise level, a candidate point to $O_i$ must exhibit a perpendicular distance to the plane above the average perpendicular distance between all plane inliers and the plane itself, plus a safe margin $\upsilon$: 

\begin{equation}
O_i = \left\{\mathbf{p}\in A''_i, d(\mathbf{p},\mathbf{\pi}_i) < \upsilon + \frac{1}{|I_i|}\sum_{\forall \mathbf{q}\in I_i}d(\mathbf{q},\mathbf{\pi}_i) \right\}.
\end{equation}

\subsubsection{Projection of obstacles in the camera plane}\label{sec:proj}
 
As described in Section~\ref{sec:pathdetector}, the original path detector uses a set of virtual agents to determine the location of the path on a stream of RGB images. The behaviors controlling these agents allow them to move from pixel to pixel, attracted by visual saliency in input images, while laying virtual pheromone on their way. The spatial distribution of accumulated virtual pheromone is taken as the estimated path location. However, the presence of visually salient obstacles may drive these agents away from the path, resulting in a mis-detection of the path. 

\begin{figure}
  \centering
  \subfigure[]{\includegraphics[width=2cm, height=1.6cm]{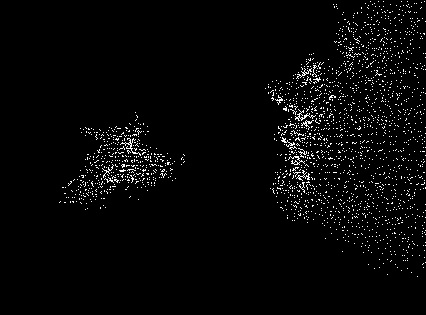}\label{fig:op_1}}\hfill
  \subfigure[]{\includegraphics[width=2cm, height=1.6cm]{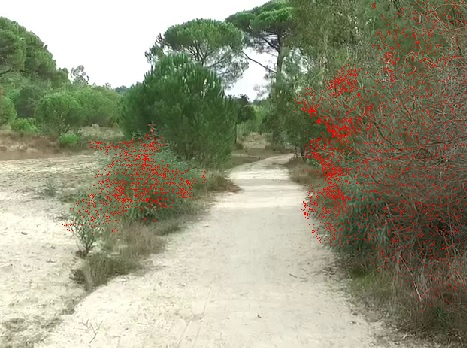}\label{fig:op_2}}\hfill
  \subfigure[]{\includegraphics[width=2cm, height=1.6cm]{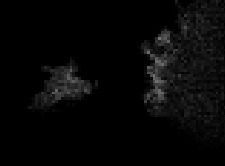}\label{fig:op_3}}\hfill
  \subfigure[]{\includegraphics[width=2cm, height=1.6cm]{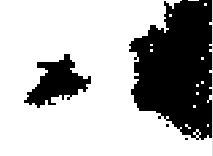}\label{fig:op_4}}\hfill
  \caption{(a) Projection of the 3D points in the plane of the camera; (b) 3D points, in red, projected on the original image; (c) Mask that indicates regions characterized as a greyscale obstacle where the lighter regions indicate a higher concentration of 3D points; (d) Binary mask indicating regions characterized as black obstacle.}\label{fig:op}
\end{figure}

To avoid being attracted by obstacles, agents need to be aware of them. Hence, at each key-frame $i$, all 3D obstacle points present in $O_i$ are projected to the camera's plane, assuming the pinhole camera model parameterized according to the off-line calibration process. The projection of all obstacle points produce a temporary $w\times h$ greyscale mask, where each pixel is graded accordingly the estimated distance to the camera sensor. 
Forthwith a binary mask is created, $M_i$, where each pixel $(x,y)$ is $M_i(x,y)=1$ if an obstacle point was projected on it and $M_i(x,y)=0$ otherwise. The resulting image (640x480) is resized (80x60), inflating obstacle regions, and passed to the path detector module. 
Being aware of these pixel-wise obstacle labels, virtual agents are able to avoid these pixels and, therefore, focus on locations of the image space more likely to belong to the path. 
Fig.~\ref{fig:op} illustrates the results obtained by the projection of 3D points in the camera's plane (\ref{fig:op_1}), where it is possible to observe that the points drawn in the original image (\ref{fig:op_2}) describe the elements above the ground plane (i.e., tall trees and shrubs), that is, the system allows to obtain a mask (\ref{fig:op_3} and \ref{fig:op_4}) where are described the possible obstacles present in the path.

\subsection{Integration with Path Detector}\label{sec:integration}

As previously stated, the binary mask $M_i$ influences the virtual agents' behavior to avoid obstacle regions in the image. This is attained by adjusting the virtual pheromone deposited by each virtual agent $a$ at each step $k$ as a function of the number of obstacle points found across its whole path. Concretely, let us assume a virtual agent $a$, deployed in the bottom row of a given conspicuity image at keyframe $i$. Let us also assume that this agent performed $m_a$ moves until it has been removed from the system, i.e., when it has reached the maximum number of moves or the maximum allowed height (row) $h_{max}$. Finally, let us assume that the original detector determined that the agent should deposit a virtual pheromone of magnitude $p_a[k]^{*}$ in the agent's position at move $k$, where $k\le m_a$. To avoid obstacle areas, it is necessary to determine how many projected obstacle points in mask $M_i$ have been crossed by the agent for all $k\in\{1,\ldots,m_a\}$:

\begin{equation}
o_a = \sum_{k=0}^{m_a} M_i(c_a[k])
\end{equation}

\noindent where $c_a[k]$ represents the pixel's coordinates (position) of agent $a$ at step $k$.

Instead of depositing the amount of pheromone defined by the original detector, $f_{a}[k]^{*}$, the agent will deposit a pheromone level whose magnitude is a non-linear function of $o_a$: 

\begin{equation}
f_a[k] = f_a[k]^* + \left(1 - \frac{o_a}{m_a}\right)^{\frac{1}{2}}.
\end{equation}

If the pheromone level calculated for the pixel visited by the virtual agent is below an empirically defined threshold, $f_a[k] < \epsilon $, no pheromone is deposited altogether. This threshold must be low enough to accommodate the presence of noise in the binary mask $M_i$.

Reducing pheromone deposition affects directly the current frame, which indirectly affects the forthcoming frames via the dynamics of the neural field, $F_i$. However, to magnify this influence, every pixels in the neural field have their activity reduced if the corresponding pixels in the obstacle map are labeled as obstacle:

\begin{equation}
F_i(x,y) \leftarrow \gamma \cdot F_i(x,y), \forall (x,y) : M_i(x,y)=1.
\end{equation}

\noindent where $\gamma$ is an empirically defined scalar stating how much the neural field should be affected, $F_i(x,y)\in [0,1]$ provides the activity level of the neural field at pixel $(x,y)$ and $M_i(x,y)\in\{0,1\}$ provides the mask value at the same pixel.

This procedure makes it possible to decrease the influence of the virtual pheromone in the regions classified as obstacles while maintaining the integrity of the neural field, i.e., without fully clearing the deposited pheromone. This is important to allow  virtual agents to recover from false obstacle detections (e.g., points belonging to the ground identified as obstacles due to mis-estimation of the ground plane).

\begin{figure}
  \centering
  \begin{minipage}[h]{1.0\linewidth}
  {\includegraphics[width=2.4cm, height=1.8cm]{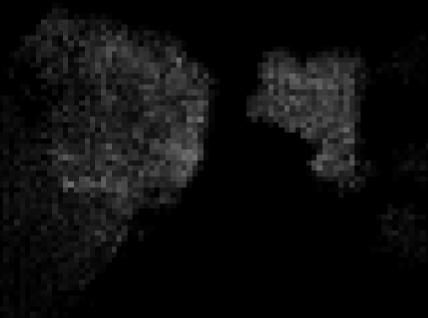}}
  {\includegraphics[width=2.4cm, height=1.8cm]{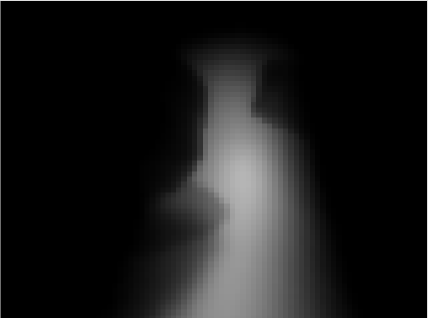}}
  {\includegraphics[width=2.4cm, height=1.8cm]{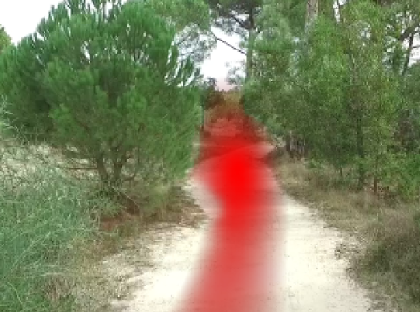}}
  \end{minipage}
  \vspace{0.1cm}
  \vspace{0.00mm}
  \begin{minipage}[h]{1.0\linewidth}
  {\includegraphics[width=2.4cm, height=1.8cm]{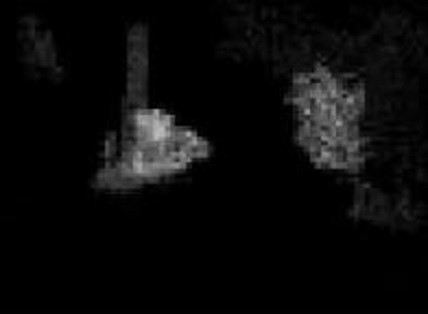}}
  {\includegraphics[width=2.4cm, height=1.8cm]{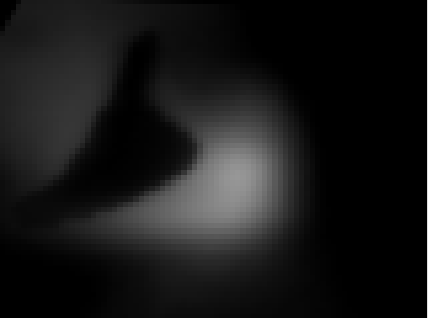}}
  {\includegraphics[width=2.4cm, height=1.8cm]{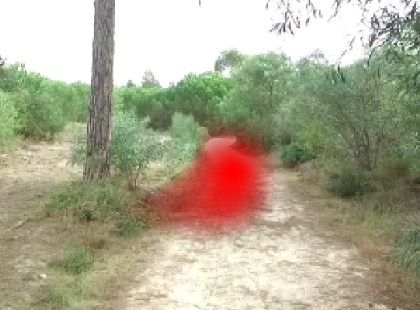}}
  \end{minipage}
  \vspace{0.1cm}
  \vspace{0.00mm}
  \begin{minipage}[h]{1.0\linewidth}
  {\includegraphics[width=2.4cm, height=1.8cm]{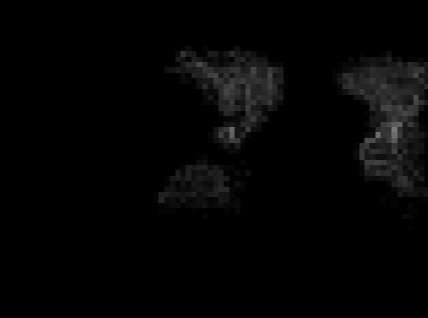}}
  {\includegraphics[width=2.4cm, height=1.8cm]{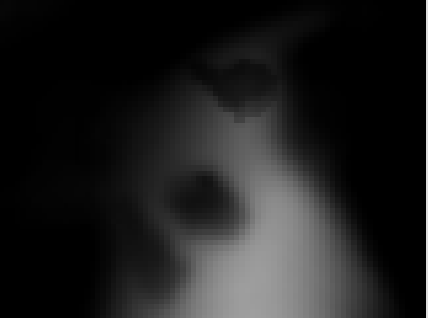}}
  {\includegraphics[width=2.4cm, height=1.8cm]{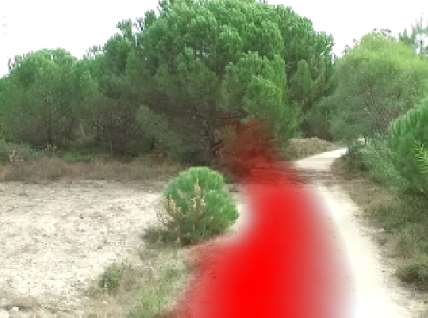}}
  \end{minipage}
  \vspace{0.1cm}
  \vspace{0.00mm}
  \caption {Process of influence on the behavior of virtual agents in order to prevent the selection of regions identified as obstacles. From left to right, obstacle mask, $M_i$, neural filter, $F_i$, and estimated path's region (highest activity blob in neural field.}\label{fig:finf}
\end{figure}  

In the original detector, virtual agents are allowed to move up to the image's row expected to match an heuristically pre-determined horizon line. This aimed at focusing the activity of virtual agents to the regions of the image below the expected horizon line, where the path is likely to appear. However, this actual horizon line depends on the camera's orientation, mostly the pitch angle, and terrain inclination. To take this into account, in each key-frame, the camera's pitch angle with respect to the estimated ground plane is determined and used to adjust the expected horizon line under a ground planar assumption.

\subsection{Validating Path Presence}\label{sec:validationsystem}

\begin{figure}[!h]
  \centering
  \subfigure[]{\includegraphics[width=2.5cm, height=1.8cm]{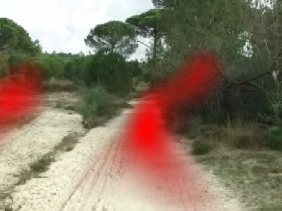}\label{fig:intercept_A}}\hfill
  \subfigure[]{\includegraphics[width=2.5cm, height=1.8cm]{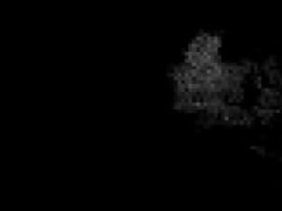}\label{fig:intercept_B}}\hfill
  \subfigure[]{\includegraphics[width=2.5cm, height=1.8cm]{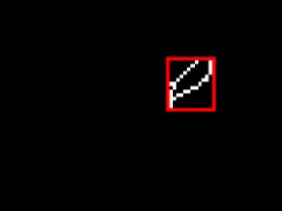}\label{fig:intercept_C}}\hfill
  \caption[Interception between the region hypothesis of the path and the regions identified as obstacles.] {Path detection validation system. (a) Detected path region marked in red. (b) Obstacle map. (c) Interception between obstacles and path region marked in white and outlined with a red rectangle.}\label{fig:intercept}\label{fig:completeSystem}
\end{figure}

Path detectors eventually fail. When that happens, the robot should be able to fall back to an alternative safe navigation strategy. Fig.~\ref{fig:completeSystem} illustrates a possible architecture in which the path detector's output is analyzed by a validation module, whose responsibility is to inform the navigation system whenever the detector's accuracy is highly compromised. 

Here we propose a validation strategy that simply checks whether there are a significant number of pixels in the input image signaled by the detector as being part of the path that are also associated to the presence of obstacles. A pixel $(x,y)$ is said to be part of the path if the the corresponding activity in the neural field is a above a given threshold, $F_i(x,y)>\nu$. The same pixel is said to be part of an obstacle if $M_i(x,y)=1$. A bounding rectangle is then fit to the pixels that meet both conditions. Fig.~\ref{fig:intercept} illustrates a situation in which the path detector is determined by this as unable to produce an accurate result.

The failure situations may occur when the path detector estimates path regions that intersect estimated obstacle regions. Hence, the validation system will be able to signal a problem when either path or obstacles are mis-detected. The latter may occur, for instance, when the SLAM method fails to reconstruct the environment or when the plane estimation procedure fails.

\section{Experiment Results}\label{cap:results}

\subsection{Experimental Setup}\label{sec:experiment}

The proposed system was developed and tested on a portable computer with a 2.20GHz Intel Core i7-4700MQ processor unit, with 16GB of RAM, running the operating system Ubuntu 14.04 LTS 64-bit (Trusty Tahr). The entire development was done using the C++ programming language on top of ROS \cite{288}, PCL \cite{Rusu_ICRA2011_PCL}, and OpenCV \cite{opencv_library}. Camera calibration was done with the corresponding ROS package, based on  \cite{zhang1999flexible,bouguet2015matlab}. The UAV used for data acquisition was a DJI Phantom 3 Advance platform \footnote{\url{https://www.dji.com/phantom-3-adv}} equipped with a HD camera used to capture 720p images at 60 fps.

After the camera calibration process, the intrinsic parameters obtained are used by LSD-SLAM in the depth map estimation process. A set of parameters is also defined in order to improve the results obtained by the LSD-SLAM in the data set used for the system's validation. Specifically, the following values were empirically defined: (a) Minimal absolute image gradient for a pixel to be considered for depth map estimation was set to 10; (b) the digital noise value in the image intensity used in its analysis was set to 5; (c) the frequency at which key-frames are obtained so as to improve the depth map in relation to the key-frame reference was set to 6; (d) the amount of new key-frames obtained in relation to the distance to the last key-frame reference was set to 16; and (e) the factor used to smooth the depth map was set to 5, i.e., the blur filter applied to the depth map to accommodate the presence of noise from the computation of the depth map. For a more detailed explanation of the values can be fount at \cite{engel2014lsd}.

The path detector uses almost the same values defined by the original detector \cite{santana2013neural}, except for the dynamically modified parameters in order to model the behavior of the virtual agents (see Section \ref{sec:integration}). Specifically, the value that defines the maximum height in the image which limits trajectories that the virtual agents can describe, $h_{max}$, is defined according to the angle of inclination of the camera relative to the ground plane, $\theta$, calculated over the execution of the system. The deposited virtual pheromone value, $f_i$, is calculated according to the number of pixels belonging to the obstacles through which the virtual agents have passed, i.e., the higher the number of pixels belonging to obstacles visited by the virtual agent, the lower the value of deposited virtual pheromone. Finally, the time filter $F_i$ is directly affected by the mask $M_i$ as a way to inhibit the virtual agents from to evaluate regions of the image where obstacles were previously identified.

\subsection{Data Set}\label{sec:data}

\begin{figure}
  \centering
    \subfigure[]{\includegraphics[width=3.6cm, height=2.4cm]{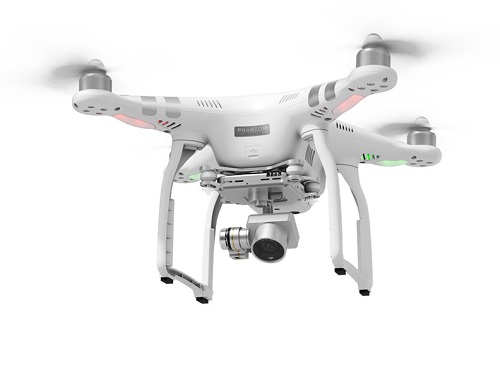}}
    \subfigure[]{\includegraphics[width=3.6cm, height=2.4cm]{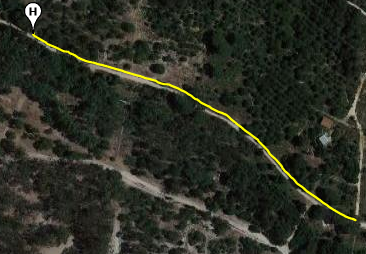}}  
    \subfigure[]{\includegraphics[width=3.6cm, height=2.4cm]{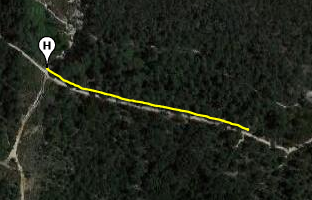}}
    \subfigure[]{\includegraphics[width=3.6cm, height=2.4cm]{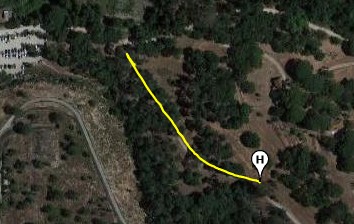}}
  \caption{UAV platform used in the acquisition of the test videos used to validate the proposed system, and examples of the paths realized by the UAV. (A) DJI Phantom 3 Advance UAV platform (image taken from the official DJI website); (B) Track taken to record Video 1; (C) Track taken to record Video 2; (D) Track taken to record Video 5; The  $H$, represents the starting point of the respective flight performed by the UAV.}\label{fig:data_path}
\end{figure}

As a way of validating the proposed system, a set of 12 videos (see Table~\ref{tab:resume}) were captured on 5 different routes at different times of the day (e.g., with the sun at its highest point), by tele-operating the UAV. After the data set of test videos has been acquired, all the processing and analysis were done offline. Fig.~\ref{fig:data_path} illustrates three of the routes made by the UAV during the data set acquisition process. A total of $\approx 43$ minutes of video were recorded, covering a total distance of $\approx 3$ kilometers, with an average distance of $\approx 300$ meters and an average time of $\approx 3.5$ minutes per video, with a mean distance from the UAV to the ground of $\approx 2.5$ meters and a speed of $\approx 1 \mbox{ms}^{-1}$. All videos were captured at a rate of 50 frames per second with a resolution of $1280\,\times\,720$ pixels, and saved with h264 video compression.

\begin{table}
\begin{center}
\centering
\begin{tabular}{|c|c|c|c|}
\hline
\makecell{Video} & \makecell{Number of \\Frames} & \makecell{Number of \\Keyframes} & \makecell{Time \\(minutes)} \\ \hline
1        & 15120            & 584                 & 5.033          \\ \hline
2        & 8150             & 283                 & 2.72           \\ \hline
3        & 13250            & 232                 & 4.42           \\ \hline
4        & 8100             & 509                 & 2.70           \\ \hline
5        & 7200             & 482                 & 2.40           \\ \hline
6        & 11300            & 509                 & 3.77           \\ \hline
7        & 13200            & 482                 & 4.40           \\ \hline
8        & 3650             & 187                 & 1.22           \\ \hline
9        & 12850            & 410                 & 4.28           \\ \hline
10       & 8550             & 337                 & 2.85           \\ \hline
11       & 10950            & 507                 & 3.65           \\ \hline
12       & 13550            & 510                 & 4.52           \\ \hline
\end{tabular}
\caption{Date set statistics.}
\label{tab:resume}
\end{center}
\end{table}

\subsection{Results}\label{sec:results}

\subsubsection{Obstacle Detection}

\begin{figure}
  \centering
    {\includegraphics[width=8cm]{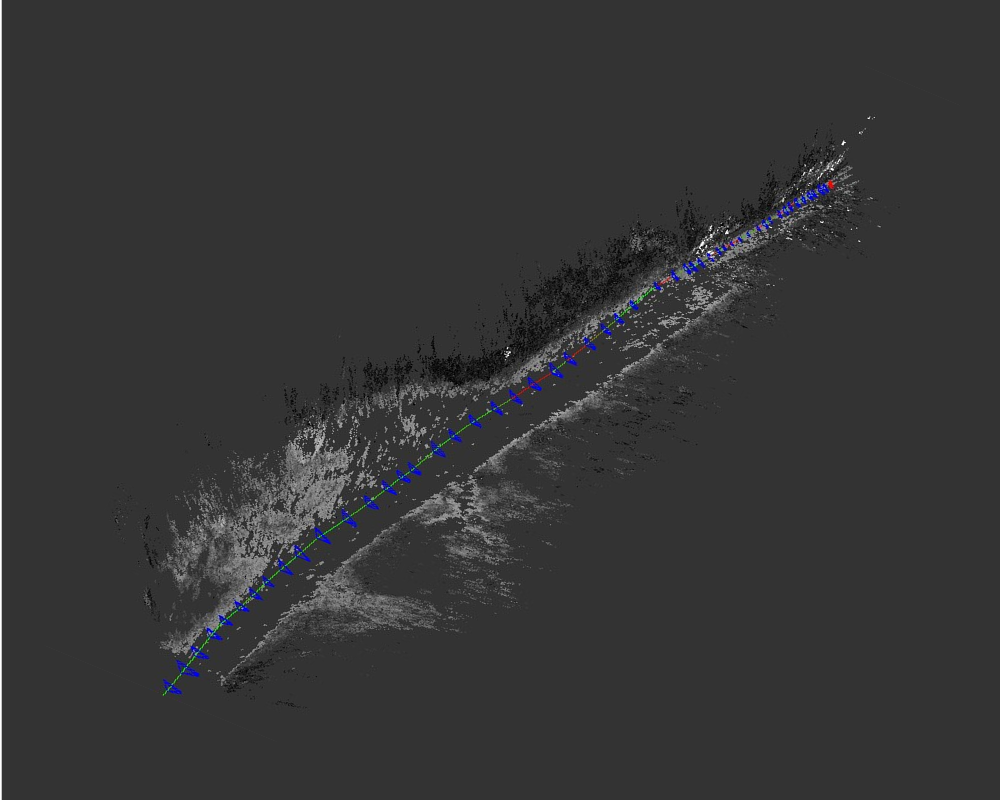}}\hfill
\caption{Example of three-dimensional reconstruction, in forest environment, obtained by LSD-SLAM without any type of filtering. The different poses of the camera are represented in blue and the path through straight segments.}\label{fig:lsdfull}
\end{figure}

\begin{figure}
  \centering
    \begin{minipage}[h]{1.0\linewidth}
      {\includegraphics[width=2cm, height=1.5cm]{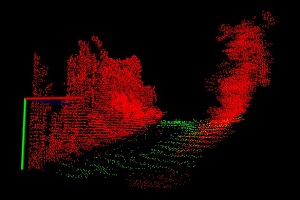}}\hfill
      {\includegraphics[width=2cm, height=1.5cm]{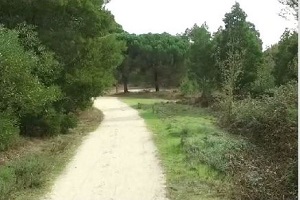}}\hfill
      {\includegraphics[width=2cm, height=1.5cm]{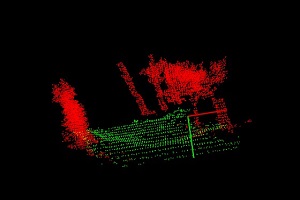}}\hfill
      {\includegraphics[width=2cm, height=1.5cm]{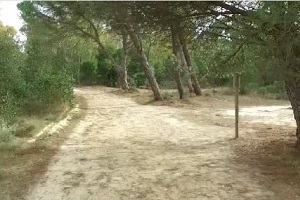}}\hfill
    \end{minipage}
    \vspace{0.08cm}
    \vspace{0.00mm}
    \begin{minipage}[h]{1.0\linewidth}
      {\includegraphics[width=2cm, height=1.5cm]{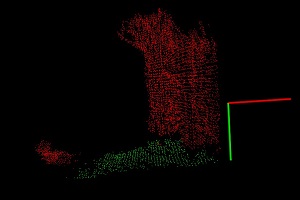}}\hfill
      {\includegraphics[width=2cm, height=1.5cm]{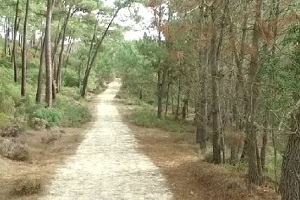}}\hfill
      {\includegraphics[width=2cm, height=1.5cm]{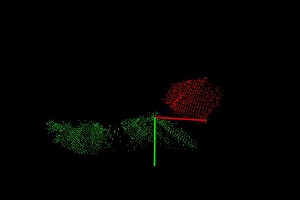}}\hfill
      {\includegraphics[width=2cm, height=1.5cm]{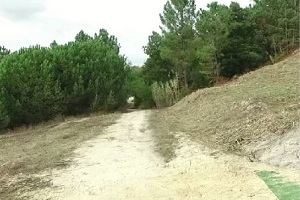}}\hfill  
    \end{minipage}
    \vspace{0.08cm}
    \vspace{0.00mm}
    \begin{minipage}[h]{1.0\linewidth}
      {\includegraphics[width=2cm, height=1.5cm]{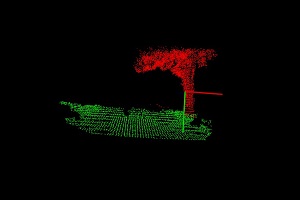}}\hfill
      {\includegraphics[width=2cm, height=1.5cm]{images/rec3_1.jpg}}\hfill
      {\includegraphics[width=2cm, height=1.5cm]{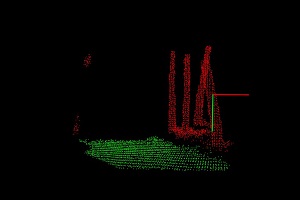}}\hfill
      {\includegraphics[width=2cm, height=1.5cm]{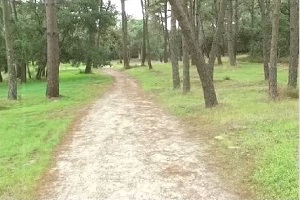}}\hfill  
    \end{minipage}
    \vspace{0.08cm}
    \vspace{0.00mm}
    \begin{minipage}[h]{1.0\linewidth}
      {\includegraphics[width=2cm, height=1.5cm]{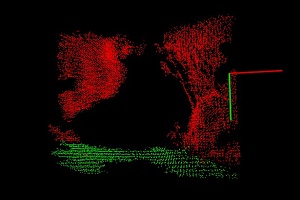}}\hfill
      {\includegraphics[width=2cm, height=1.5cm]{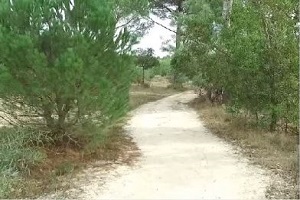}}\hfill
      {\includegraphics[width=2cm, height=1.5cm]{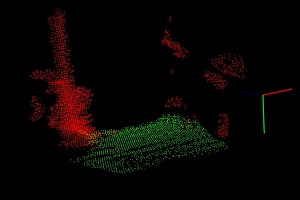}}\hfill
      {\includegraphics[width=2cm, height=1.5cm]{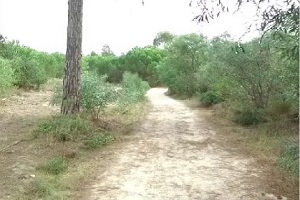}}\hfill  
    \end{minipage}
    \vspace{0.08cm}
    \vspace{0.00mm}
    \begin{minipage}[h]{1.0\linewidth}
      {\includegraphics[width=2cm, height=1.5cm]{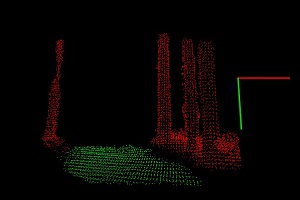}}\hfill
      {\includegraphics[width=2cm, height=1.5cm]{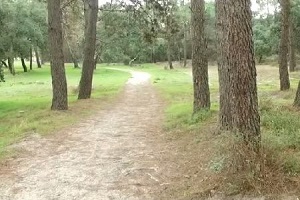}}\hfill
      {\includegraphics[width=2cm, height=1.5cm]{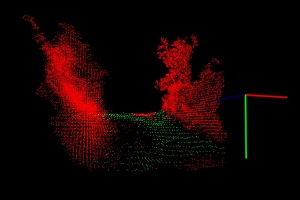}}\hfill
      {\includegraphics[width=2cm, height=1.5cm]{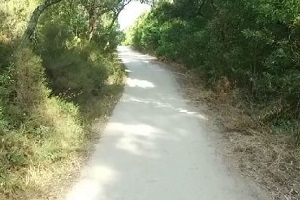}}\hfill  
    \end{minipage}
    \vspace{0.08cm}
    \vspace{0.00mm}
\caption[Sub-set of results that demonstrates the three-dimensional reconstruction of the environment.] {Sub-set of typical results obtained with the set of test data, demonstrating the three-dimensional reconstruction of the environment described in front of the robot ($ \approx 8 \mbox{m})$. Images with three-dimensional reconstruction, first and third columns, illustrate the clouds of points obtained where, the red dots describe the obstacles and the green dots describe the ground plane.}\label{fig:recfilter}
\end{figure}

Figure~\ref{fig:lsdfull} presents a typical result concerning the 3D reconstruction obtained directly from the LSD-SLAM, when applied to the data set used during tests. The product of the 3D reconstruction is composed of a cloud of accumulated points without any kind of treatment, which may contain millions of points. After processing the accumulated point cloud (see Section~\ref{sec:segmentation}), a filtered point cloud can be obtained. Figure~\ref{fig:recfilter} illustrates the 3D reconstruction of the different elements that are close to the robot properly segmented as obstacle and ground points by the method described in Section~\ref{sec:segmentation}. These results show the ability of a monocular visual SLAM system to produce adequate representations for obstacle detection in the vicinities of the path.

\begin{figure}
  \centering
  	\begin{minipage}[h]{1.0\linewidth}
      {\includegraphics[width=2cm, height=1.5cm]{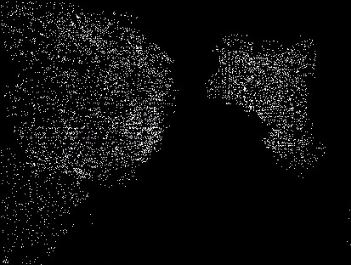}}\hfill
      {\includegraphics[width=2cm, height=1.5cm]{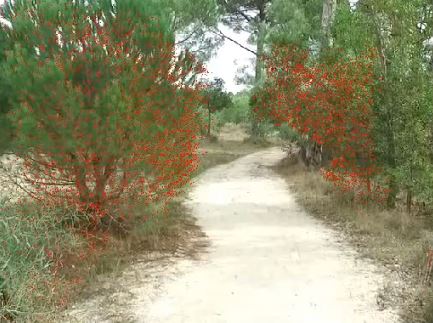}}\hfill
      {\includegraphics[width=2cm, height=1.5cm]{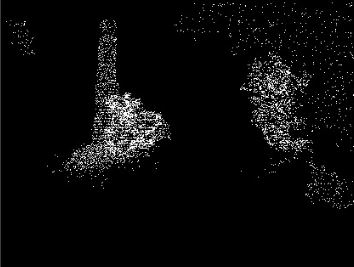}}\hfill
      {\includegraphics[width=2cm, height=1.5cm]{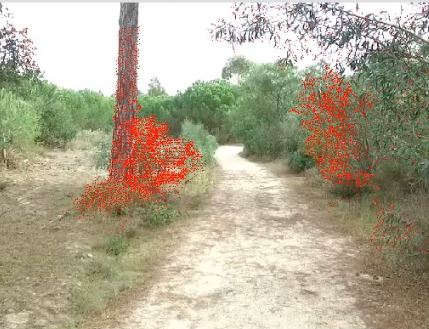}}\hfill  
    \end{minipage}
    \vspace{0.08cm}
    \vspace{0.00mm}
    \begin{minipage}[h]{1.0\linewidth}
      {\includegraphics[width=2cm, height=1.5cm]{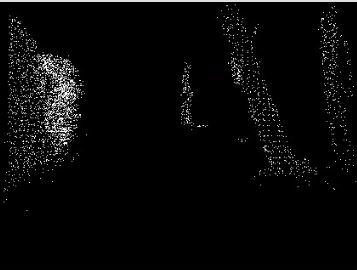}}\hfill
      {\includegraphics[width=2cm, height=1.5cm]{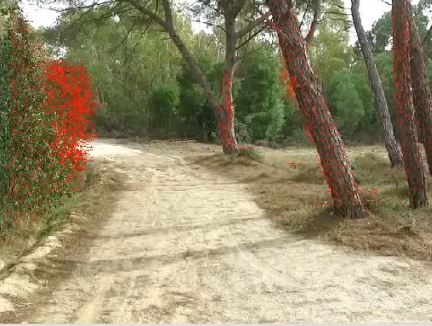}}\hfill
      {\includegraphics[width=2cm, height=1.5cm]{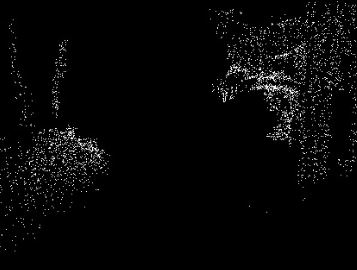}}\hfill
      {\includegraphics[width=2cm, height=1.5cm]{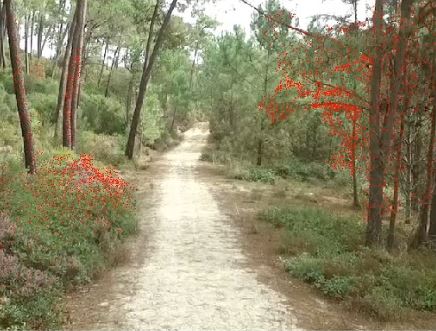}}\hfill  
    \end{minipage}
    \vspace{0.08cm}
    \vspace{0.00mm}
    \begin{minipage}[h]{1.0\linewidth}
      {\includegraphics[width=2cm, height=1.5cm]{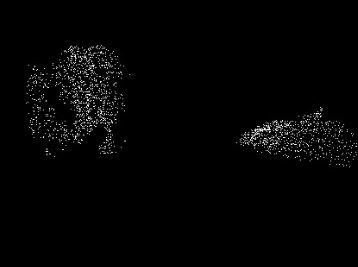}}\hfill
      {\includegraphics[width=2cm, height=1.5cm]{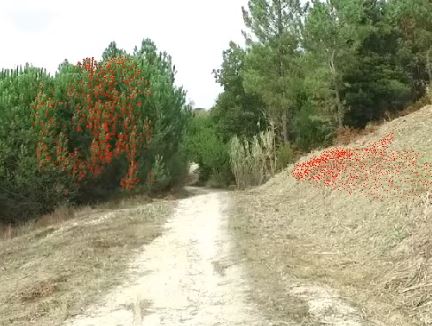}}\hfill
      {\includegraphics[width=2cm, height=1.5cm]{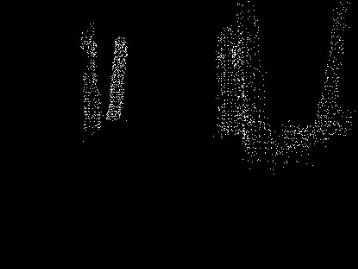}}\hfill
      {\includegraphics[width=2cm, height=1.5cm]{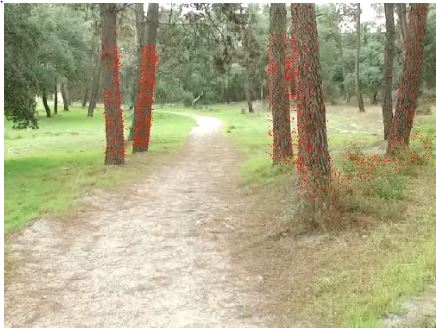}}\hfill  
    \end{minipage}
    \vspace{0.08cm}
    \vspace{0.00mm}
    \begin{minipage}[h]{1.0\linewidth}
      {\includegraphics[width=2cm, height=1.5cm]{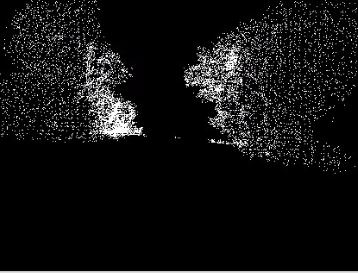}}\hfill
      {\includegraphics[width=2cm, height=1.5cm]{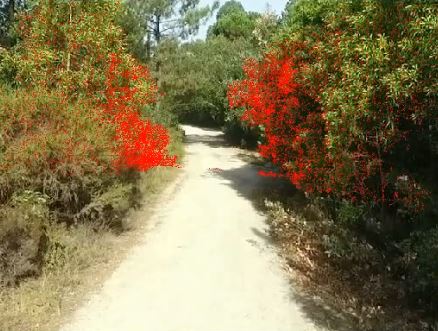}}\hfill
      {\includegraphics[width=2cm, height=1.5cm]{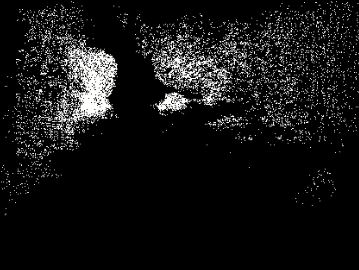}}\hfill
      {\includegraphics[width=2cm, height=1.5cm]{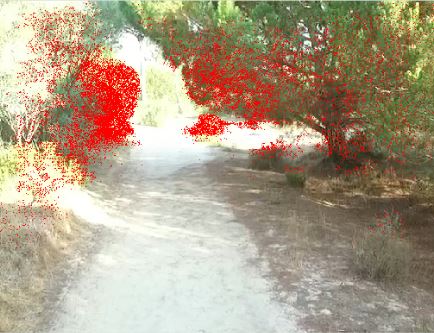}}\hfill  
    \end{minipage}
    \vspace{0.08cm}
    \vspace{0.00mm}    
\caption[Typical sub-set of results that demonstrates the correct identification of the near obstacles of the robot.] {Sub-set typical results obtained with the set of test data that demonstrates a correct identification of the near obstacles of the robot, present in its field of vision. The results are presented in pairs of images, where, on the left is the projection of the points obtained by the three-dimensional reconstruction, in the plane of the camera, and to the right the obstacles described in the original image as red dots.}\label{fig:obsfi}
\end{figure}

Fig.~\ref{fig:obsfi} depicts typical results referring to the projection of detected obstacles onto the image plane of key-frames to build a binary mask, i.e., an obstacle map. The figure also shows the obstacle map overlaid on the RGB image of the corresponding key-frames. These results show that the obstacles are properly mapped to the RGB image in which virtual agents operate, thus be readily available as an additional sensory input for them to avoid obstacle populated image regions. 

\subsection{Path Detection and Tracking}\label{sec:r4}

\begin{figure}
  \centering
     \begin{minipage}[h]{1.0\linewidth}
       {\includegraphics[width=2cm, height=1.5cm]{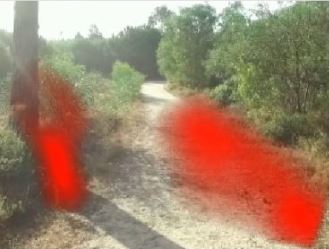}}\hfill
       {\includegraphics[width=2cm, height=1.5cm]{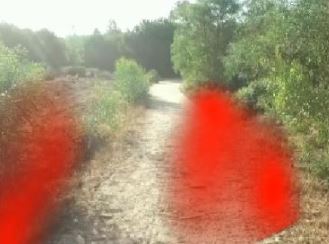}}\hfill
       {\includegraphics[width=2cm, height=1.5cm]{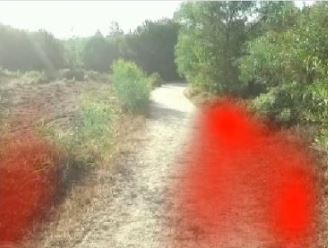}}\hfill
       {\includegraphics[width=2cm, height=1.5cm]{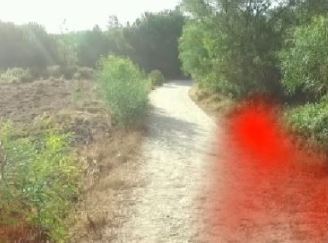}}\hfill
    \end{minipage}
    \vspace{0.08cm}
    \vspace{0.00mm}
    \begin{minipage}[h]{1.0\linewidth}
      {\includegraphics[width=2cm, height=1.5cm]{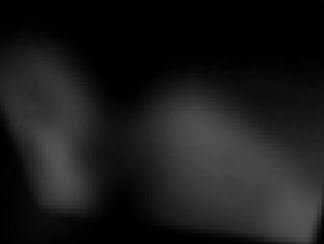}}\hfill
      {\includegraphics[width=2cm, height=1.5cm]{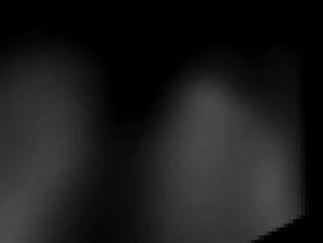}}\hfill
      {\includegraphics[width=2cm, height=1.5cm]{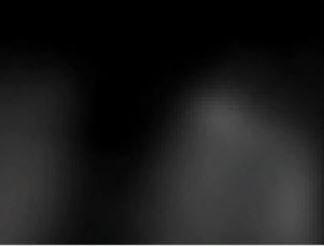}}\hfill
      {\includegraphics[width=2cm, height=1.5cm]{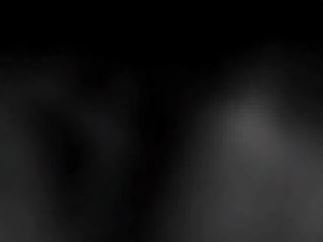}}\hfill  
    \end{minipage}
    \vspace{0.08cm}
    \vspace{0.00mm}
    \begin{minipage}[h]{1.0\linewidth}
      {\includegraphics[width=2cm, height=1.5cm]{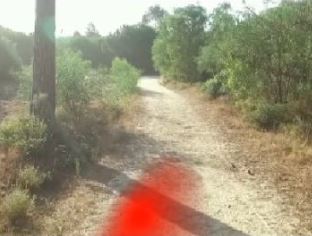}}\hfill
      {\includegraphics[width=2cm, height=1.5cm]{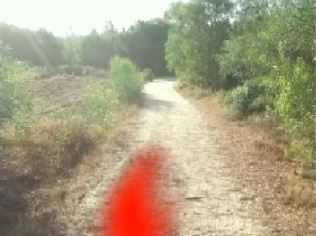}}\hfill
      {\includegraphics[width=2cm, height=1.5cm]{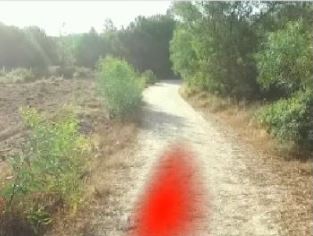}}\hfill
      {\includegraphics[width=2cm, height=1.5cm]{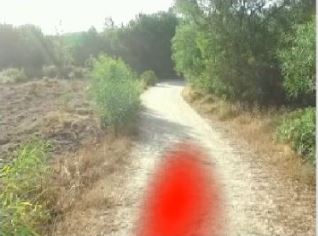}}\hfill
    \end{minipage}
    \vspace{0.08cm}
    \vspace{0.00mm}
    \begin{minipage}[h]{1.0\linewidth}
      {\includegraphics[width=2cm, height=1.5cm]{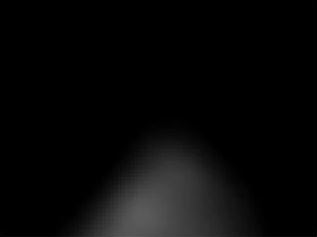}}\hfill
      {\includegraphics[width=2cm, height=1.5cm]{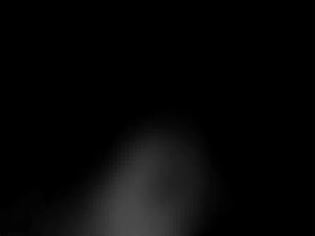}}\hfill
      {\includegraphics[width=2cm, height=1.5cm]{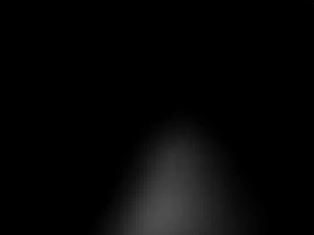}}\hfill
      {\includegraphics[width=2cm, height=1.5cm]{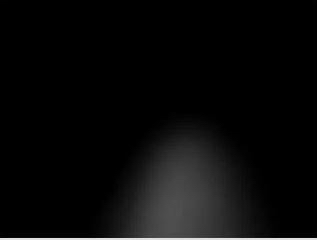}}\hfill  
    \end{minipage}
    \vspace{0.08cm}
    \vspace{0.00mm}
    \begin{minipage}[h]{1.0\linewidth}
      {\includegraphics[width=2cm, height=1.5cm]{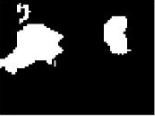}}\hfill
      {\includegraphics[width=2cm, height=1.5cm]{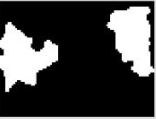}}\hfill
      {\includegraphics[width=2cm, height=1.5cm]{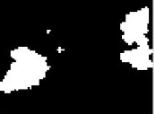}}\hfill
      {\includegraphics[width=2cm, height=1.5cm]{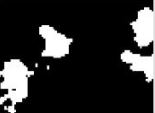}}\hfill  
    \end{minipage}
    \vspace{0.08cm}
    \vspace{0.00mm}
    
\caption[Images taken from tests performed in Video 6.] {Images taken from tests performed in Video 6. The first line illustrates the result obtained by the original detector, the third line illustrates the result obtained by the proposed system, the second and fourth lines illustrate the time filter, $\text{NF}$, which shows activity of the virtual agents relative to the virtual pheromone deposit through a red spot and line five illustrates the detection of obstacles by the proposed system. The set of results was obtained by running both systems on the keyframes generated by the LSD-SLAM.}\label{fig:comp}
\end{figure}

\begin{figure}
  \centering
    \begin{minipage}[h]{1.0\linewidth}
      {\includegraphics[width=1.6cm, height=1.2cm]{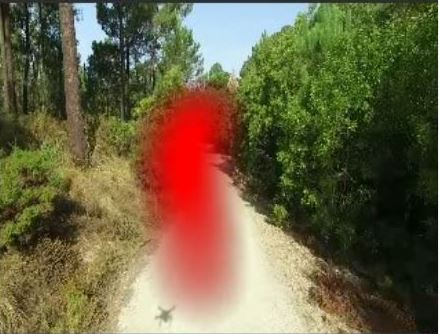}}\hfill
      {\includegraphics[width=1.6cm, height=1.2cm]{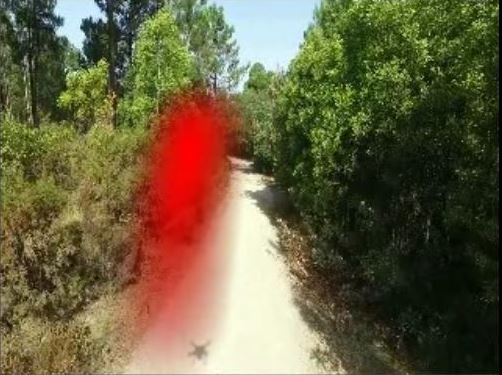}}\hfill
      {\includegraphics[width=1.6cm, height=1.2cm]{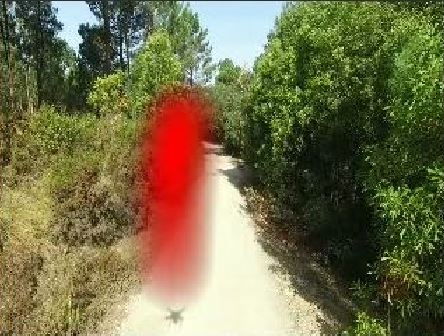}}\hfill
      {\includegraphics[width=1.6cm, height=1.2cm]{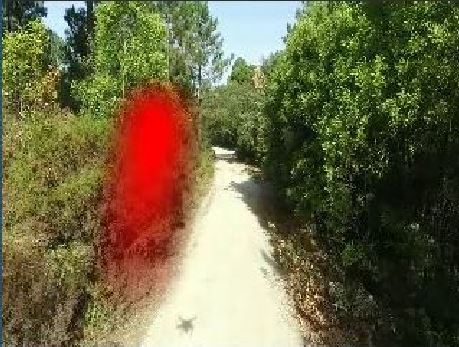}}\hfill
      {\includegraphics[width=1.6cm, height=1.2cm]{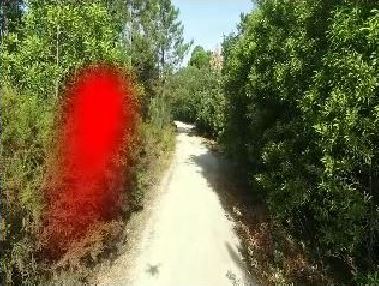}}\hfill
    \end{minipage}
    \vspace{0.05cm}
    \vspace{0.00mm}
    \begin{minipage}[h]{1.0\linewidth}
      {\includegraphics[width=1.6cm, height=1.2cm]{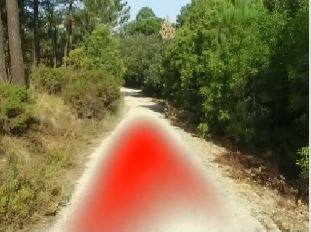}}\hfill
      {\includegraphics[width=1.6cm, height=1.2cm]{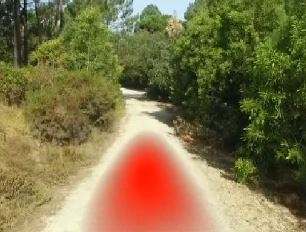}}\hfill
      {\includegraphics[width=1.6cm, height=1.2cm]{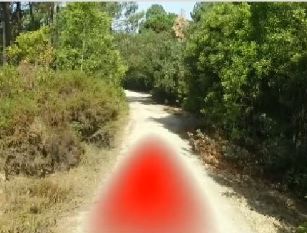}}\hfill
      {\includegraphics[width=1.6cm, height=1.2cm]{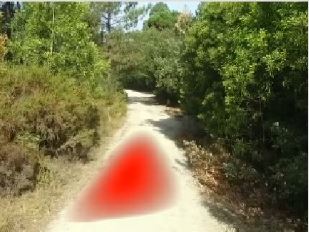}}\hfill
      {\includegraphics[width=1.6cm, height=1.2cm]{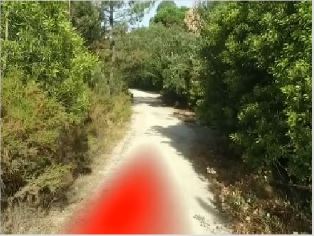}}\hfill
    \end{minipage}
    \vspace{0.05cm}
    \vspace{0.00mm}
\caption[Images of the result obtained by the original detector in a fault situation.] {Images of the result obtained by the original detector in a fault situation in the test performed on the complete video sequence of Video 12. In the first line is the result obtained by the original detector is shown, in the second line is shown the result obtained by the proposed system.}\label{fig:comp2}
\end{figure}

\begin{figure}
  \centering
    \subfigure[]{\includegraphics[width=2cm, height=1.5cm]{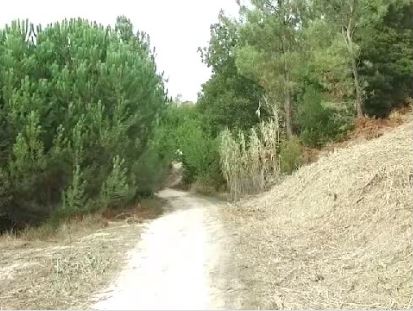}}\hfill
    \subfigure[]{\includegraphics[width=2cm, height=1.5cm]{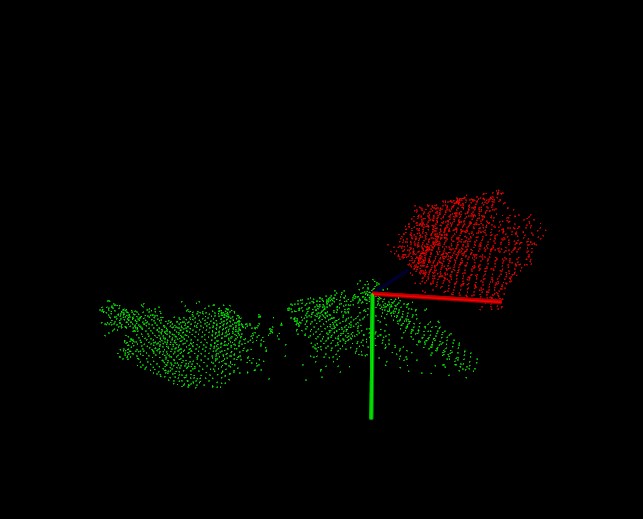}}\hfill
    \subfigure[]{\includegraphics[width=2cm, height=1.5cm]{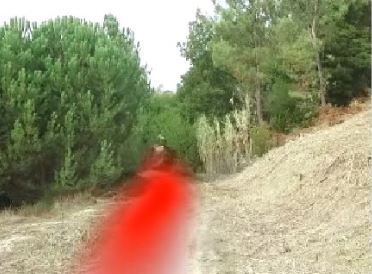}}\hfill
    \subfigure[]{\includegraphics[width=2cm, height=1.5cm]{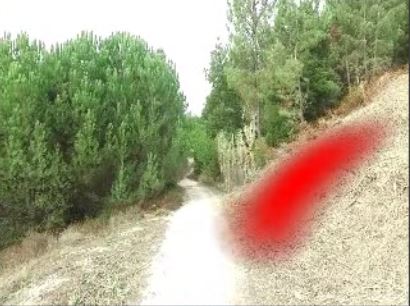}}\hfill
\caption[Case of original detector failure following Video 6 images.] {(A) Image analyzed; (B) Three-dimensional information where the green dots represent the ground plane and the red dots represent the region to the right of the highest path; (C) Region of the path identified by the proposed system; (D) Region hypothesis of the path identified by the original detector.}\label{fig:comp3}
\end{figure}

Fig.~\ref{fig:finf} it is possible to observe the regions characterized as an obstacle in the data set and their influence on the temporal filter used to identify the region of the path hypothesis. Specifically, it is noted that virtual agents tend to reject regions of the image identified as obstacles, thereby helping to produce more correct detections. Figure~\ref{fig:comp} compares the proposed system and the original detector in the same situation of Video 6 under equal conditions, that is, where only the set of keyframes available from LSD-SLAM were used.
It can be seen that the original detector is influenced by the light conditions that make the region to the left of the image (i.e., vegetation) more prominent than the path, then being selected as the region of the path hypothesis. On the other hand, the proposed system is able to identify this same region as an obstacle (line 5 of Figure~\ref{fig:comp}) and thus prevent virtual pheromones being deposited in this region (line four of Figure~\ref{fig:comp}), avoiding that it is chosen as the hypothesis region of the path. In order to make a fair comparison to the original detector, it was tested in the complete image sequence of Video 12 (not only the keyframes), where it is possible to observe, through Figure~\ref{fig:comp2}, a result similar to that obtained previously, i.e., the original detector continues to fail to identify the region of the image where the path is located. Another example can be seen in Figure~\ref{fig:comp3}, referring to the test performed with Video 3, where the proposed system is able to determine that the region identified by the original detector as belonging to the path is much high that the actual path.


Since the result obtained by the proposed system, like the original detector, consists of the identification of a hypothesis region that must be contained in a part of the path and not a complete segmentation of the path, the chosen quantitative evaluation method follows a similar approach to the one used to evaluate the original detector \cite{santana2013neural}. Specifically, the track is considered correctly identified if the red spot representing the regions of the temporal filter ($\text{NF}$) which record intensities above $80\%$ of the maximum intensity, is located within the limits of the path and its spatial distribution is aligned with the direction of the path. In situations where there is ambiguity generated by the occurrence of two red spots in the resulting image, that is, when two hypothesis regions are identified by the system for the path at the same time, the result must be chosen as the result produced by the system that has a higher intensity in the filter temporal.

\begin{center}
\begin{table}
\begin{small}
\centering
\caption{Percentage of correctly analyzed keyframes by the original detector and by the herein proposed system. Mean and respective standard deviation obtained in the five tests performed by video.}
\label{tab:results}
\begin{tabular}{cccc}
\hline
\makecell{Video \\ number} & \makecell{number of \\ Keyframes} & \makecell{Original \\Detector [\%]  } & \makecell{Proposed \\System [\%] } \\ 
\hline
\hline
1  & $585 \pm 32$ & $89.98 \pm 1.19$ & $98.35 \pm 1.05$ \\
\hline
2  & $283 \pm 27$ & $99.33 \pm 0.92$  & $100.0  \pm 0.00$ \\
\hline
3  & $470 \pm 15$ & $77.67 \pm 7.06$ & $97.77 \pm 0.52$ \\
\hline
4  & $320 \pm 20$ & $95.79 \pm 3.04 $ & $95.61 \pm 1.02$ \\
\hline
5  & $235 \pm 12$ & $86.54 \pm 5.96$ & $96.00 \pm 3.83$ \\
\hline
6  & $501 \pm 25$ & $59.19 \pm 21.52$ & $98.64 \pm 1.07$ \\
\hline
7  & $482 \pm 31$ & $94.94 \pm 0.78$ & $97.91 \pm 1.16$ \\
\hline
8  & $187 \pm 19$ & $94.05 \pm 4.77$ & $97.60 \pm 2.94$ \\
\hline
9  & $411 \pm 22$ & $93.18 \pm 2.04$ & $96.52 \pm 1.02$ \\
\hline
10  & $337 \pm 23$ & $99.81 \pm 0.39$ & $100.0  \pm 0.00$ \\
\hline
11  & $507 \pm 36$ & $95.79 \pm 2.02$ & $97.41 \pm 1.14$ \\
\hline
12 & $510 \pm 11$ & $96.96 \pm 3.04$ & $97.77 \pm 2.23$ \\ 
\hline
\hline
\multicolumn{2}{c}{Overall detection rate [$\%$]:}  & $\mathbf{90.30 \pm  4.22}$ & $\mathbf{97.78 \pm 1.09}$ \\
\end{tabular}
\end{small}
\end{table}
\end{center}

\begin{figure}
  \centering
    \begin{minipage}[h]{1.0\linewidth}
      {\includegraphics[width=2cm, height=1.5cm]{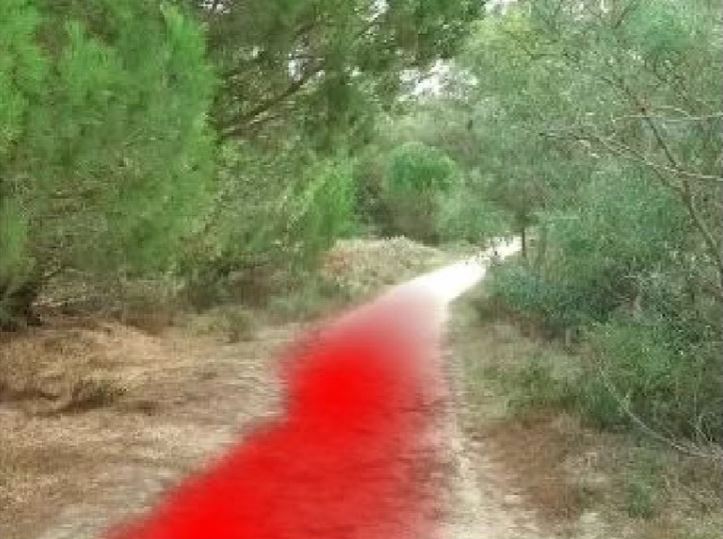}}\hfill
      {\includegraphics[width=2cm, height=1.5cm]{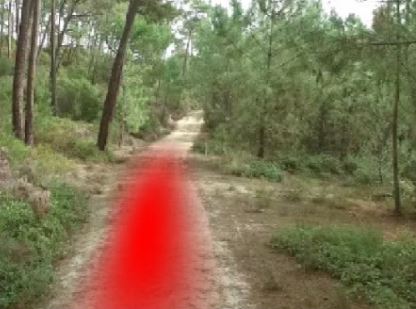}}\hfill
      {\includegraphics[width=2cm, height=1.5cm]{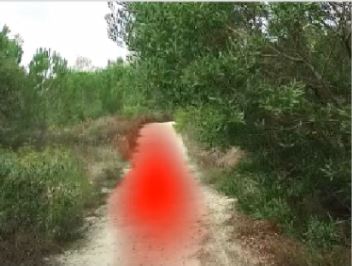}}\hfill
      {\includegraphics[width=2cm, height=1.5cm]{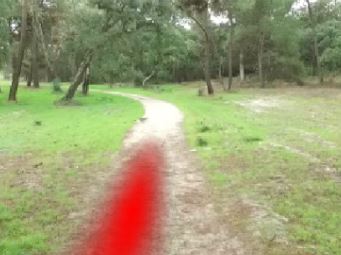}}\hfill  
    \end{minipage}
    \vspace{0.05cm}
    \vspace{0.00mm}
    \begin{minipage}[h]{1.0\linewidth}
      {\includegraphics[width=2cm, height=1.5cm]{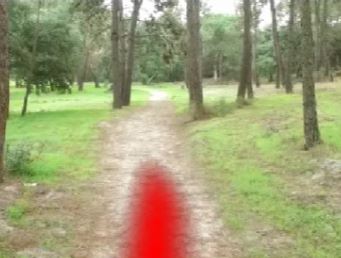}}\hfill
      {\includegraphics[width=2cm, height=1.5cm]{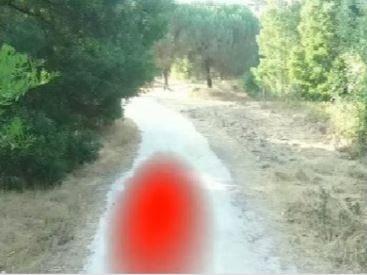}}\hfill
      {\includegraphics[width=2cm, height=1.5cm]{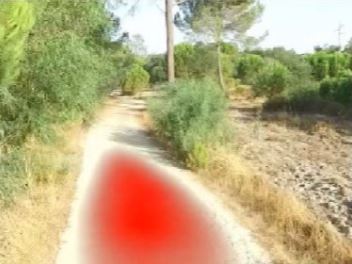}}\hfill
      {\includegraphics[width=2cm, height=1.5cm]{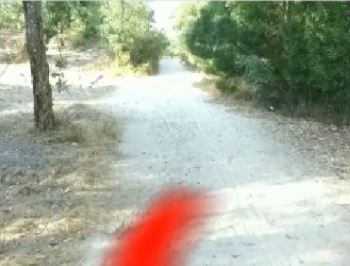}}\hfill  
    \end{minipage}
    \vspace{0.05cm}
    \vspace{0.00mm}
    \begin{minipage}[h]{1.0\linewidth}
      {\includegraphics[width=2cm, height=1.5cm]{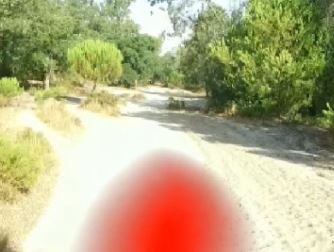}}\hfill
      {\includegraphics[width=2cm, height=1.5cm]{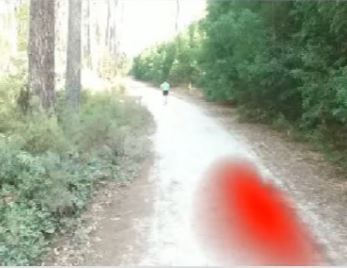}}\hfill
      {\includegraphics[width=2cm, height=1.5cm]{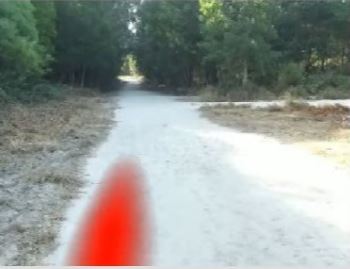}}\hfill
      {\includegraphics[width=2cm, height=1.5cm]{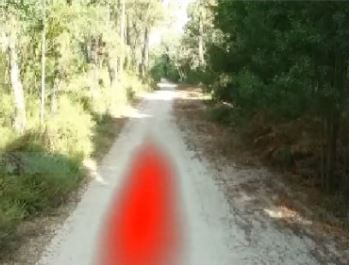}}\hfill  
    \end{minipage}
    \vspace{0.05cm}
    \vspace{0.00mm}
    
\caption[Result obtained in the identification of the hypothesis region of the path by video.] {Result obtained in the identification of the region hypothesis of the path by video. Left to right and top to bottom of Video 1 to Video 12.}\label{fig:resulttotal}
\end{figure}

Analyzing the results obtained shows that the original detector obtains a success rate of $90.30\%$ with a standard deviation of $4.22\%$ in the data set used. It is confirmed that the original detector shows to fail more frequently when influenced by the presence of distractors in the regions surrounding the path, which is aggravated by the fact that in these situations the detector learns the appearance of these distractors as being the appearance of the path, causing it to propagate the fault over and, inevitably, guide the robot to regions potentially occupied by obstacles. On the other hand, the proposed system has a smaller standard deviation of $1.09\%$, with a success rate in the identification of the path hypothesis of $97.78\%$. With a positive gain of $7.86\,\%$ success than the original detector, accompanied by a lower standard deviation, the proposed system proves the benefits of including 3D information in modulating the behavior of the virtual agents presented by the original detector, thus validating the fourth hypothesis (R4). Figure~\ref{fig:resulttotal} illustrates a set of typical results obtained by the proposed system, where it is possible to observe the correct identification of the path hypothesis.

\subsection{Validation System}\label{sec:r5}

Although the path detector proposed in this dissertation shows greater robustness relative to the original as a result of the integration of three-dimensional information, as discussed in Section~\ref{sec:r4}, there are factors that may cause it to fail. For example, sudden changes in the level of illumination may lead to failures in the visual projection method, abrupt movements of the camera in poor lighting conditions can lead to optical flow calculation failures and abrupt changes in the appearance of the path may lead to the system learns the wrong appearance, leading it to deviate from the correct identification of the region's path hypothesis. The use of three-dimensional environmental information has been shown to mitigate the occurrence of these situations; however, in extreme situations as identified above, the influence exerted by volumetric information on virtual agents may not be sufficient to prevent regions of the image occupied by obstacles being classified as a region of the path. The validation module described in Section~\ref{sec:validationsystem} aims at informing the navigation system when these situations occur, allowing it to disregard the detector.

\begin{figure}
  \centering
  	\begin{minipage}[h]{1.0\linewidth}
      {\includegraphics[width=2cm, height=1.5cm]{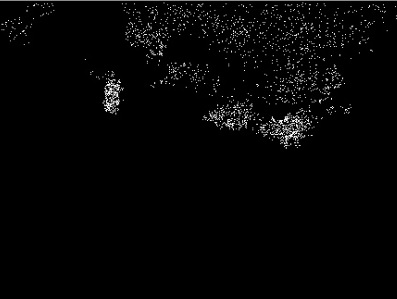}}\hfill
      {\includegraphics[width=2cm, height=1.5cm]{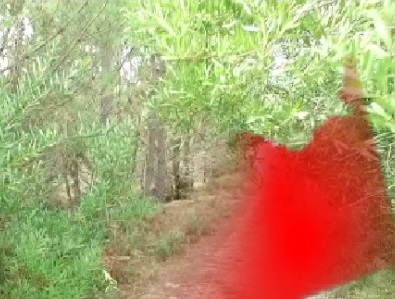}}\hfill
      {\includegraphics[width=2cm, height=1.5cm]{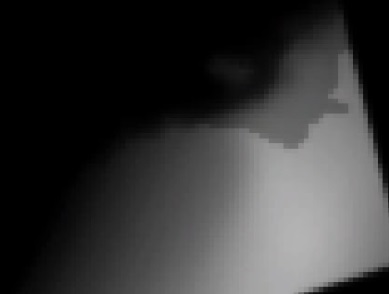}}\hfill
      {\includegraphics[width=2cm, height=1.5cm]{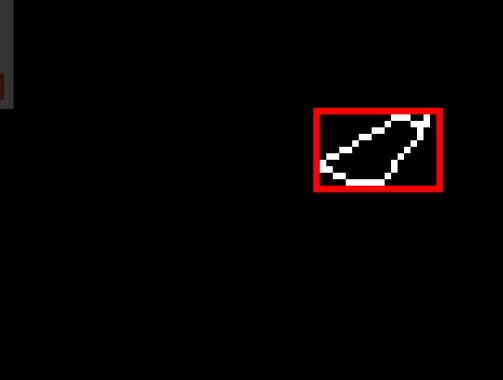}}\hfill
    \end{minipage}
    \vspace{0.05cm}
    \vspace{0.00mm}
    \begin{minipage}[h]{1.0\linewidth}
      {\includegraphics[width=2cm, height=1.5cm]{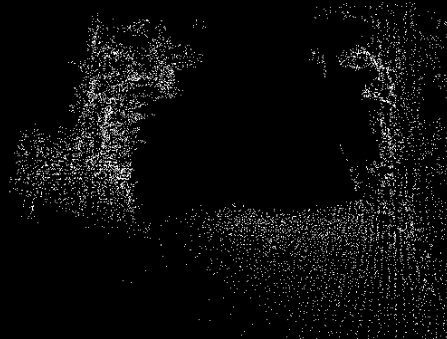}}\hfill
      {\includegraphics[width=2cm, height=1.5cm]{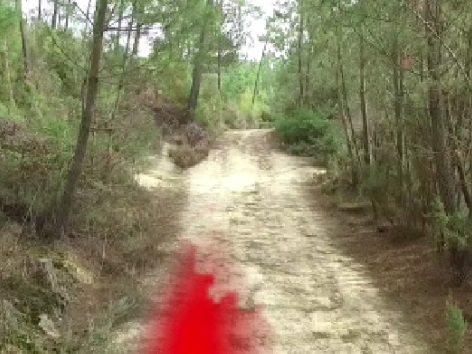}}\hfill
      {\includegraphics[width=2cm, height=1.5cm]{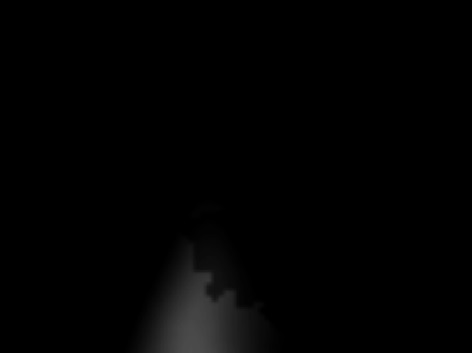}}\hfill
      {\includegraphics[width=2cm, height=1.5cm]{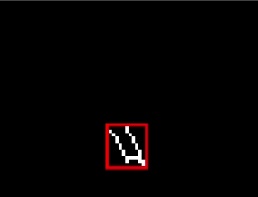}}\hfill
    \end{minipage}
    \vspace{0.05cm}
    \vspace{0.00mm}
\caption[Sub-set of typical results that demonstrate the operation of the Validator module.] {Sub-set of typical results that demonstrate the operation of the Validator module. The first line illustrates the identification of a fault caused by the calculation of the optical flow. The second line illustrates the identification of a fault caused by an erroneous signaling of obstacles in the region where the path is located. The first column illustrates the projection of the obstacle presence information to the camera plane, the second column illustrates the result of the track detector, the third column illustrates the activity of the neuronal field, $ NF $, and the fourth column illustrates the result obtained by the validator module.}\label{fig:resval}
\end{figure}

The first line of Figure~\ref{fig:resval} depicts a situation in which  there is an optical flow failure due to rapid camera movement in conjunction with poor light conditions, which leads to a spreading of field activity neuronal interaction that induces the track detector to erroneously classify a region occupied by obstacles. As a result, the validator module signals the occurrence of a failure. In poor 3D reconstruction situations, the plane estimation algorithm may fail, causing the system to erroneously report obstacles in the region where the path is located. In this case, the validator module signals how the discrepancy between the position of the obstacles detected (wrongly above the path) and the position of the path estimated by the path detector, as it is possible to observe in the second line of Figure~\ref{fig:resval}. 

\subsection{Limitations}\label{sec:failure}

The proposed system demonstrated, through the results obtained, to be able to identify regions potentially dangerous to the robot and to successfully change the behavior of the original detector to produce a more robust system for detection and tracking of paths. However, the proposed system still presents some challenges that can be solved in the sequence of the work carried out in this article. One of the main challenges identified is the computational cost required for the acquisition of three-dimensional environmental information through a monocular camera using the LSD-SLAM software package. Directly related to computational cost is the time required by LSD-SLAM to generate depth maps, averaging $1.154 \pm 0.27$ seconds. However, with the advancement of computing power and the emergence of high-performance systems such as Nvidia TX2\footnote{Nvidia TX2: \url{http://www.nvidia.com/object/embedded-systems-dev-kits-modules.html}}, which allows a greater parallelism in the execution of tasks, together with the improvement that has been observed in the optimization of SLAM techniques could mitigate this issue in a short time. The proposed system was designed to be modular, through the ROS architecture, however this modularity also brings a computational cost associated with the exchange of messages between the different modules, thus as an alternative solution the integration of all the code in a single software package more optimized would eliminate the cost associated with the exchange of messages, to the detriment of the modularity of the system.

In addition to the time required between acquisition of depth maps, the proposed system requires sufficient translation of the camera so that the LSD-SLAM can estimate the three-dimensional structure of the environment.

\begin{figure}
  \centering
  	\begin{minipage}[h]{1.0\linewidth}
      {\includegraphics[width=2cm, height=1.5cm]{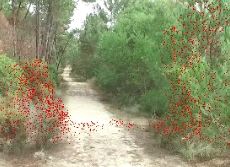}}\hfill
      {\includegraphics[width=2cm, height=1.5cm]{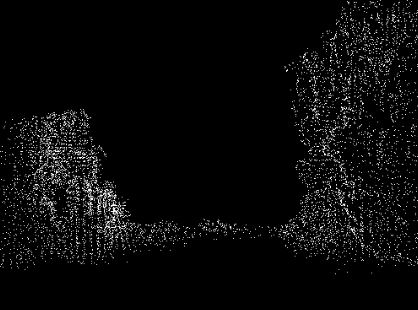}}\hfill
      {\includegraphics[width=2cm, height=1.5cm]{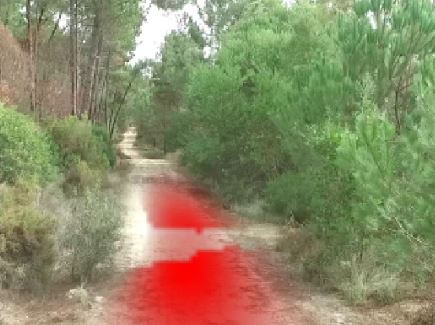}}\hfill
      {\includegraphics[width=2cm, height=1.5cm]{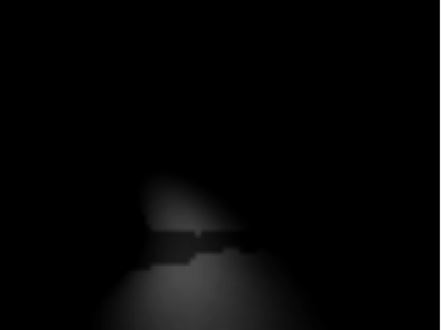}}\hfill
    \end{minipage}
    \vspace{0.05cm}
    \vspace{0.00mm}
  	\begin{minipage}[h]{1.0\linewidth}
      {\includegraphics[width=2cm, height=1.5cm]{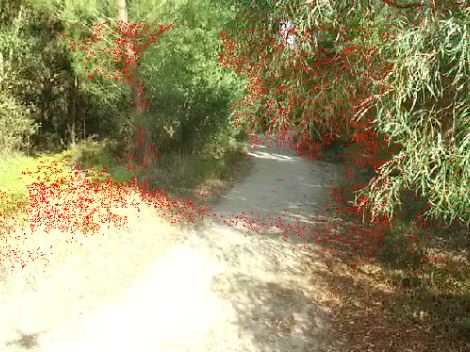}}\hfill
      {\includegraphics[width=2cm, height=1.5cm]{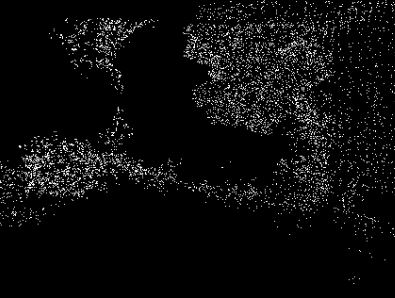}}\hfill
      {\includegraphics[width=2cm, height=1.5cm]{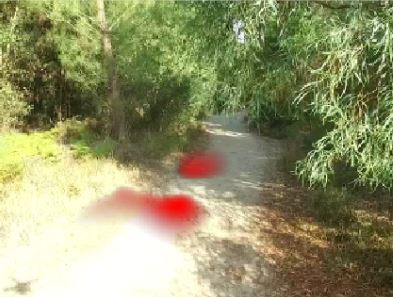}}\hfill
      {\includegraphics[width=2cm, height=1.5cm]{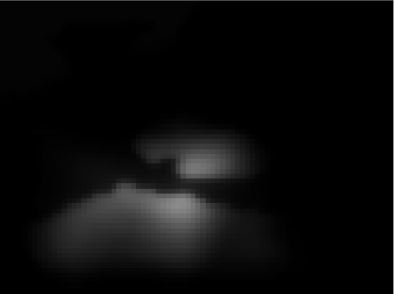}}\hfill
    \end{minipage}
    \vspace{0.05cm}
    \vspace{0.00mm}        
\caption[Failure cases identified during the tests.] {Failure cases identified during testing, where first column illustrates overlapping of three-dimensional information in the original image, the second column illustrates the projection of three-dimensional information in the camera plane, the third column illustrates the overlap of the temporal filter,  $\text{NF}$, in the original image, and the fourth column illustrates the temporal filter, $\text{NF}$.}\label{fig:fail1}
\end{figure}

Thus, a correct estimation of the ground plane, where the path is described, is extremely important, in order to segment the obstacles that are close to the robot. However, due to path irregularities, which are sometimes observed in forest environments, the assumption of planar terrain may become fragile, sometimes resulting in false positives in the detection of obstacles. In this way, the parameter that defines what is an obstacle as a function of height to the ground plane becomes very relevant in more irregular paths. Figure~\ref{fig:fail1} illustrates two situations where the value set for the height is not high enough, which promotes the identification of regions belonging to the path as an obstacle. Consequently, virtual agents are influenced by the presence of an obstacle that does not actually exist, which causes an interruption in the temporal filter (see the fourth column of Figure~\ref{fig:fail1}). However, the proposed system proved to be able to recover from such situations in the following images, due to the exploitation capacity of the virtual agents and to the way the time filter is affected by the obstacle mask, $ m_{obs}$, as discussed in Section~\ref{sec:integration}.

In Section~\ref{sec:r5} are identified fault situations, which typically occur in extreme situations, where factors such as sudden change of luminosity or rapid camera movements do not allow the proposed system to correctly identify the region of the path. The validator module was able to mitigate these situations through its identification which allows, for example, to notify the navigation system when the degree of certainty of the correct identification of the region of the path hypothesis. In the absence of three-dimensional information the proposed system behaves and fails in situations where the original detector fails. However, as soon as the 3D information is retrieved, the detector tends to recover, which is not the case for the original detector.

\section{Conclusions and Future Work}\label{cap:cfw}

This paper presented an extension to a previous work \cite{santana2013neural} for detection and tracking of tracks from monocular cameras. The main extension refers to the use small sized unmanned aerial vehicles that compute volumetric information obtained from a visual slam system, reducing the sensitivity of the detector in the presence of distractors.

The experimental results obtained show that the proposed system is able to obtain better results than the original detector in the used test dataset. Specifically, the proposed system obtains a success rate, in the identification of the path hypothesis region, of $97.78\%$, compared to the $90.30\%$ obtained by the original detector. The ability of the proposed system to estimate the angle of inclination of the camera relative to the ground (i.e., ground plane) was also evaluated, where it is possible to conclude, through a qualitative analysis of the results, that the obtained value shows a good approximation in relation to the expected value, for the observations made during the UAV flight. However, as a way of validating these results a quantitative analysis is necessary, which should be carried out in future work. Another interesting result is that the 3D reconstruction of the distracting elements present in the robot's field of vision, even though it is not complete (i.e., sparse 3D point cloud), allows to identify obstacles with some resolution, which can also be explored in future work, for example, to identify the type of elements present in the path or surrounding regions through the analysis of their volumetric signature.

However, despite the good results obtained, the proposed system presents a considerable computational cost due to the acquisition of the 3D information of the environment, which consequently requires the control at low speed of the UAV (e.g., less than $1\mbox{ms}^{-1}$). On the other hand, the increase in the processing capacity of the current mobile devices, the emergence of high performance integrated SOC systems and the great progress that has been made in recent years in the development and optimization of SLAM techniques for monocular cameras  \cite{younes2016survey}, can allow systems such as the one proposed in this work to process a greater amount of information and consequently allow the control of UAV at higher speeds and also increase the robustness in the control at low speeds. Another possible solution to this problem is the optimization of the proposed detector to allow a complete analysis of the sequence of images, instead of only analyzing key-frames. This improvement should therefore be analyzed and developed in future work.

\section{Acknowledgements}\label{cap:ack}

The research for this paper was partially supported by Introsys S.A. which provided all the necessary tools and infrastructure. We also thank Magno Guedes, Research Engineer at Introsys S.A., and Raquel Caldeira, Head of R\&D Department at Introsys S.A., for the support and comments during the development of this work.

\bibliographystyle{spmpsci} 
\bibliography{references}

\end{document}